\documentclass[10pt,twocolumn,letterpaper]{article}

\usepackage[pagenumbers]{cvpr} 

\definecolor{cvprblue}{rgb}{0.21,0.49,0.74}
\usepackage[pagebackref,breaklinks,colorlinks,allcolors=cvprblue]{hyperref}
\usepackage{float}
\usepackage{placeins}
\usepackage{siunitx}
\usepackage{makecell}
\usepackage{pifont}
\usepackage{subcaption}

\newcommand{\bfqty}[2]{\text{\bfseries\SI{#1}{#2}}}

\def\paperID{5} 
\def\confName{CVPR}
\def\confYear{2025}

\title{Towards Ball Spin and Trajectory Analysis in Table Tennis Broadcast Videos via Physically Grounded Synthetic-to-Real Transfer}

\author{Daniel Kienzle$^{1\text{,}2}$ \quad Robin Schön$^{1}$ \quad Rainer Lienhart$^{1}$ \quad Shin'Ichi Satoh$^{2\text{,}3}$ \vspace{0.3em} \\
{\normalsize $^1$University of Augsburg, Germany, {\tt\small \{firstname.lastname\}@uni-a.de}} \quad \\
{\normalsize $^2$ National Institute of Informatics, Japan}
\quad \\
{\normalsize $^3$ University of Tokyo, Japan}
}

\begin{document}
\maketitle
\begin{abstract}
Analyzing a player's technique in table tennis requires knowledge of the ball's 3D trajectory and spin. 
While, the spin is not directly observable in standard broadcasting videos, we show that it can be inferred from the ball's trajectory in the video. 
We present a novel method to infer the initial spin and 3D trajectory from the corresponding 2D trajectory in a video. 
Without ground truth labels for broadcast videos, we train a neural network solely on synthetic data.
Due to the choice of our input data representation, physically correct synthetic training data, and using targeted augmentations, the network naturally generalizes to real data. 
Notably, these simple techniques are sufficient to achieve generalization.
No real data at all is required for training.
To the best of our knowledge, we are the first to present a method for spin and trajectory prediction in simple monocular broadcast videos, achieving an accuracy of \bfqty{92.0}{\%} in spin classification and a 2D reprojection error of \bfqty{0.19}{\%} of the image diagonal.
\end{abstract}    

\vspace{-0.2cm}
\section{Introduction}
\label{sec:intro}
Computer vision is widely used across various sports to enhance athlete performance, analyze opponents' strategies \cite{general1,general2}, and assist referees with automated decision-making such as goal-line technology in soccer or line-calling in tennis \cite{hawkeye,goalcontrol}. 
Additionally, it provides valuable insights for sports broadcasting. \\[0.5ex]
\noindent While player movement analysis is crucial in many sports, table tennis can rely mostly on ball trajectories. Notably, spin plays a key role in understanding gameplay, making its estimation essential for detailed performance analysis. 
This work especially focuses on predicting ball spin besides its 3D trajectory from standard table tennis broadcast videos. \\[0.5ex]
\noindent We propose a learning-based method to estimate the ball's spin and 3D trajectory, enabling a comprehensive analysis of its kinematics. 
Broadcast video analysis is challenging due to low frame rates, small ball size, and motion blur. 
Since no public ground truth exists, we train our model exclusively on synthetic data. 
Instead of using raw visual data, we extract the ball's and table's 2D positions in the video frames, providing an abstract yet effective game-state representation.
This representation in addition to physically correct synthetic data and targeted augmentations enables our model to generalize to real data. \\[0.5ex]
\begin{figure}[t]
    \centering
    \includegraphics[width=\linewidth]{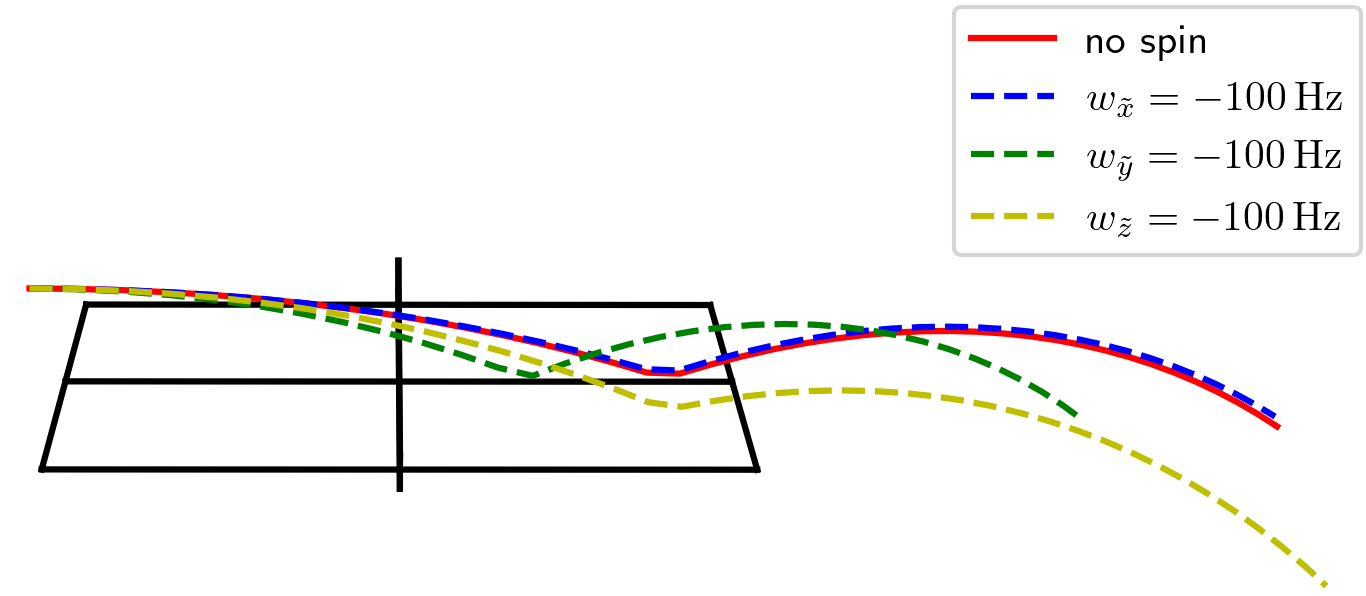}
    \vspace{-0.2cm}
    \caption{Simulated trajectory of the ball in the image plane under the influence of different spin components $\omega_{\tilde{x}}$, $\omega_{\tilde{y}}$, and $\omega_{\tilde{z}}$.
    }
    \vspace{-0.3cm}
    \label{img:spincomponents}
\end{figure}
\noindent Our main contributions are:
\begin{itemize}
    \item \textbf{Spin \& trajectory estimation}: We introduce a novel approach for analyzing table tennis gameplay by predicting ball spin and 3D trajectory. To our knowledge, this is the first spin estimation method applicable to standard monocular broadcast videos.
    \item \textbf{Synthetic-to-real generalization}: Our model generalizes to real-world data despite being trained solely on synthetic data. Three simple techniques are sufficient to achieve excellent generalization: Using a smart input representation, utilizing a physics-based simulation for the synthetic data, and implementing suitable augmentations.
    \item \textbf{Synthetic train data \& simulation pipeline}: We generate a large dataset of synthetic trajectories and publish both the dataset and simulation pipeline.
    \item \textbf{Annotated real-world test data}: We have manually annotated broadcast videos to evaluate real-world generalization. We make this dataset publicly available.
\end{itemize}

\section{Related Work}
\label{sec:relwork}

\vspace{-0.1cm}
\paragraph{3D Trajectory Estimation}
Ball trajectory analysis is crucial across various sports, with many methods relying on expensive multi-camera setups for triangulation \cite{triangulation1,triangulation2,triangulation3,triangulation4}. 
While highly accurate, this is costly and the data is often inaccessible. 
In contrast, we focus on monocular broadcast videos, making our method widely applicable. \\[0.5ex]
\noindent Some works estimate a ball's 3D position in individual frames using observed size or height cues \cite{BallLocalizationSingleImages,BallLocalizationSingleImages2,BallLocalizationSingleImagesPhysics}. 
\cite{TableTennisBallSize} applies this to table tennis but requires extremely high-frame-rate footage. In total, single-frame visual cues are generally insufficient for a precise localization. 
Instead, we uplift 2D detections over time to infer the full 3D trajectory.\\[0.5ex]
\noindent Several methods fit physical models to observed 2D trajectories for 3D estimation in sports like badminton \cite{MonoTrackBadminton}, volleyball \cite{3DRegressionVolleyball}, and basketball \cite{3DRegressionBasketball}. 
These typically require camera calibration and estimated turning points, which can introduce errors. 
Rather than performing an optimization for each trajectory, we train a neural network once to directly predict the 3D trajectory in an end-to-end approach, improving stability by performing camera calibration only indirectly and eliminating turning point estimates. \\[0.5ex]
\noindent Deep learning enables direct 2D-to-3D mapping. 
Similar to our work, \cite{synthnet} uses synthetic 2D trajectories for training, predicting the initial 3D position of a tennis ball before applying a physics model. 
However, their approach overlooks ball spin and complex interactions like bouncing and the Magnus effect. 
In contrast, our physics simulation provides more realistic training data, and we predict the full 3D trajectory instead of the initial conditions to avoid cumulative errors in motion equation solutions.

\vspace{-0.4cm}
\paragraph{Spin Estimation}
While spin is crucial in table tennis, it is not directly observable in standard broadcasts. 
To our knowledge, we present the first method for spin estimation from monocular broadcast videos. \\[0.5ex]
Existing approaches often rely on specialized hardware: Event cameras \cite{TableTennisSpinEventCamera,TableTennisSpinEventCamera2}, high-speed cameras detecting ball markings \cite{SpinDOE}, or multi-camera setups recognizing logos \cite{MultiCameraLogoPrediction}. 
While effective, these methods require controlled environments or ball modifications, making them impractical for real-world broadcast analysis. 
Our method overcomes these limitations by estimating spin from standard video footage without additional equipment.

\vspace{-0.4cm}
\paragraph{Physics in Computer Vision}
Previous works have e.g. used physics-based losses for 3D localization \cite{BallLocalizationSingleImagesPhysics}, simulated ball trajectories for trajectory completion \cite{blackbox}, and physical models to estimate landing positions in golf \cite{PhysicsGolfBall}.
In table tennis, \cite{TableTennisPINN} applies physics-informed neural networks (PINNs) \cite{PINNs} to multi-camera 3D trajectory data to extract ball motion properties, and \cite{trajectoryprediction1,trajectoryprediction2} predict the future trajectory based on previous 3D measurements.
Moreover, \cite{AchievingHumanLevel} trains a robotic player entirely on synthetic data.
Our approach leverages the physical simulation environment from \cite{AchievingHumanLevel} to generate synthetic data, enabling our model to generalize to real-world broadcast videos.

\begin{figure*}[t]
    \centering
    \includegraphics[width=0.92\linewidth]{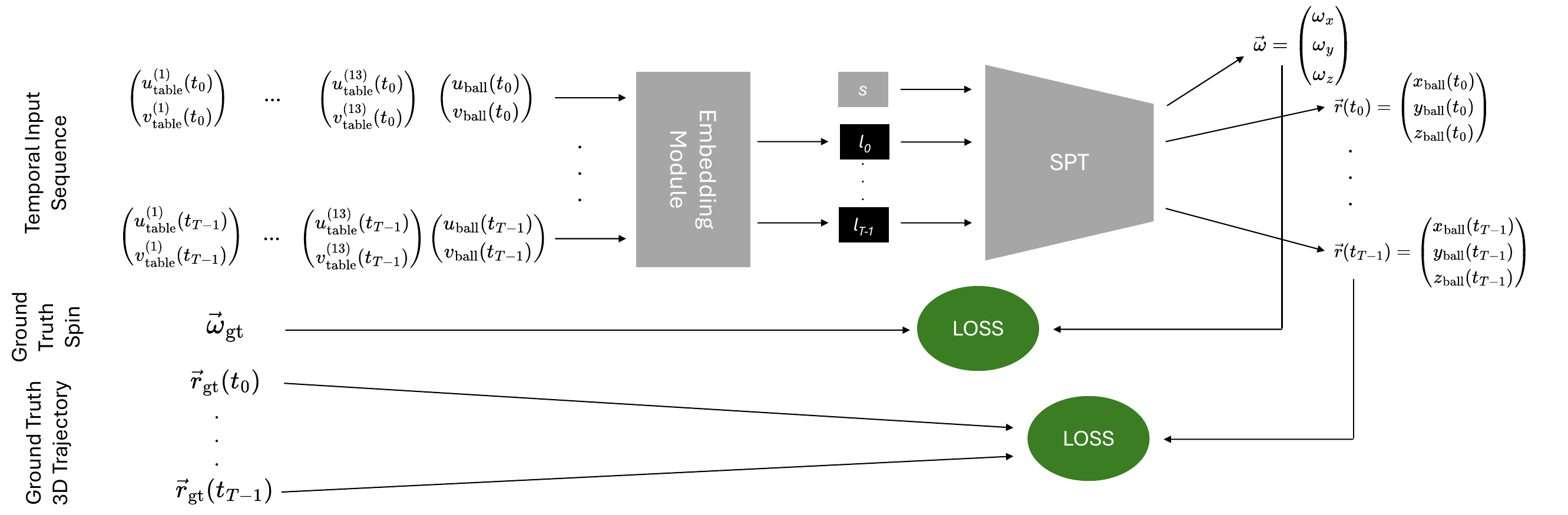}
    \caption{Overview of our pipeline. For each time step $t_i$, ball coordinates and table keypoints are embedded to generate location tokens $l_i$. A learnable spin token $s$ is prepended, and the SPT processes the sequence $\{s, l_0, ..., l_{T-1}\}$. The SPT predicts the initial spin $\vec{\omega}$ at $t_0$ and the sequence of 3D ball positions $\{\vec{r}(t_0), ..., \vec{r}(t_{T-1})\}$. The predicted spin and trajectory are supervised using separate loss terms, ensuring accurate learning of both components.
    }
    \label{fig:pipeline}
    \vspace{-0.4cm}
\end{figure*}

\section{Problem Description}
\label{sec:problem}
To effectively analyze a player's technique in table tennis, it is essential to understand the ball's trajectory and the initial spin imparted upon contact with the paddle.
This section first outlines the setup of our method and discusses the key design choices.
We then introduce appropriate coordinate systems for analyzing both trajectory and spin.
Finally, we examine how spin influences the trajectory and identify which spin components can be accurately observed.

\subsection{Main Goal}
We define the ball's 3D position at each video frame $i$ at time $t_i$ as $\vec{r}(t_i)$.
Similarly, we denote the ball's initial spin as $\vec{\omega} = \vec{\omega}(t_0)$.
Our model predicts both $\vec{r}(t_i)$ for each $t_i$ and $\vec{\omega}$ once for the full trajectory. \\[0.5ex]
\noindent The input to our model consists of the observed 2D pixel coordinates of the ball for each time $t_i$.
We employ an end-to-end approach that does not require additional information such as camera calibration.
To facilitate this, we also extract the 2D image coordinates of 13 key table points per frame, since they contain valuable information about the orientation of the camera.
Thus, the input to our model comprises the ball's 2D trajectory along with the 2D positions of these table points.
Figure \ref{img:tablekeypoints} in the supplementary material illustrates these key table points in a single frame. \\[0.5ex]
\noindent Rather than using raw video frames, our approach relies on a smart data representation based on extracted 2D coordinates of the ball and table points.
This is a widely used technique in computer vision, commonly applied in fields such as 3D pose estimation \cite{hpe1,hpe2,hpe3}, action recognition \cite{actionrecognition1,actionrecognition2}, and sign language translation \cite{signlanguage1}.
Since 2D keypoint extraction is a well-established field \cite{hpe2d3,hpe2d1,hpe2d2}, we do not further elaborate on this step.
In this paper we assume that the 2D keypoint extraction is already performed and focus on the subsequent steps. \\[0.5ex]
A significant advantage of this smart data representation is that it enables our model to focus on the most relevant information while avoiding distractions from irrelevant visual details, which could lead to overfitting and reduced generalization.
This way, the model's attention is naturally directed towards the key information, which is crucial for the success of our method.
Furthermore, while generating realistic synthetic video data is challenging, it is relatively straightforward to simulate accurate synthetic trajectories and spin using physics-based models.
Consequently, this data representation enables us to solely train the model on synthetic data and still achieve good generalization on real data. \\[0.5ex]
\noindent A major challenge when working with broadcast footage is the absence of ground-truth annotations, making it impossible to train a model directly on real data.
To address this, we generate a large dataset of physically accurate synthetic trajectories for training.
To ensure a smooth transition from synthetic to real data, we employ an advanced physics simulation engine that accurately models the ball's interactions with the table.
For this purpose, we use the MuJoCo simulation engine \cite{mujoco}, utilizing the same simulation parameters as in \cite{AchievingHumanLevel}.
Since \cite{AchievingHumanLevel} demonstrated that synthetic-to-real transfer is feasible in table tennis simulations, we are confident that their setup aligns well with our requirements.

\subsection{Predictions and Coordinate Systems}
\label{sec:coordsystems}
\vspace{-0.2cm}
\begin{figure}[th]
    \centering
    \includegraphics[width=0.82\linewidth]{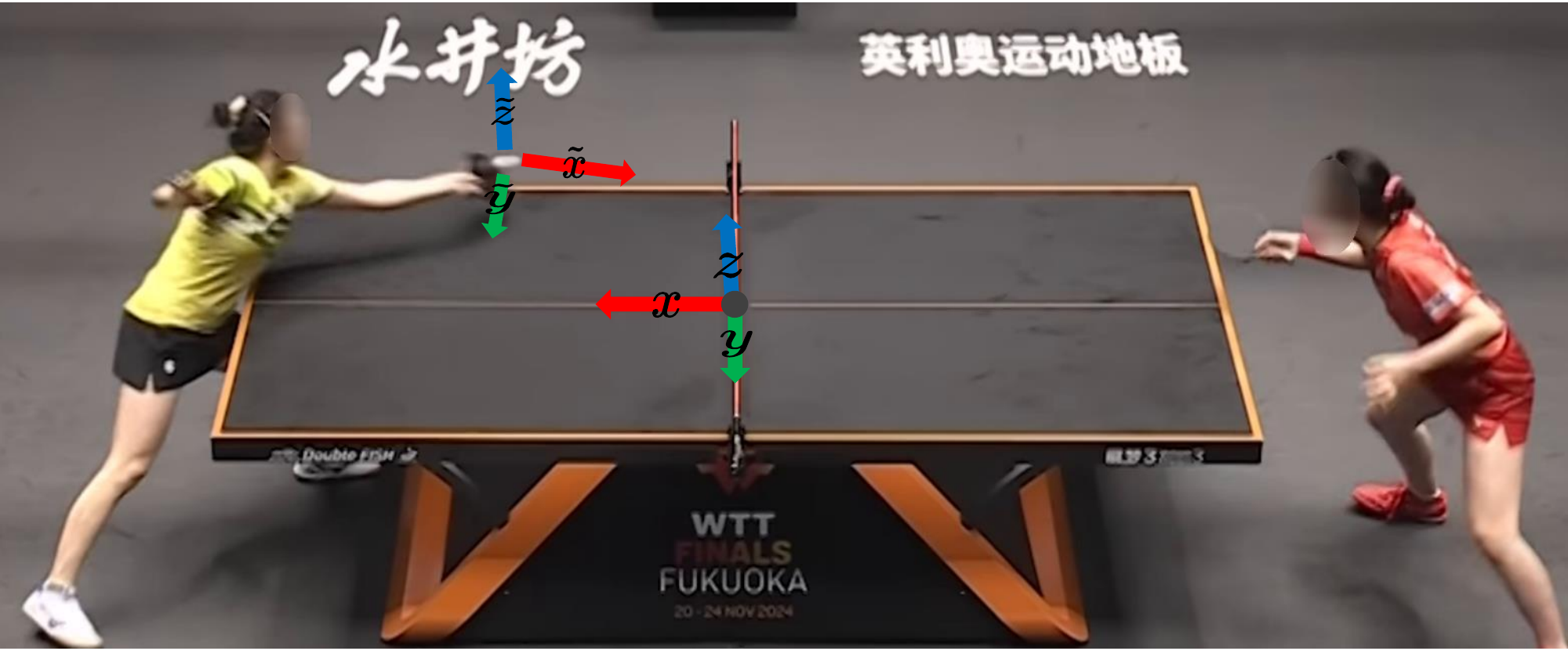}
    \vspace{-0.2cm}
    \caption{World coordinate system ($x$, $y$, $z$) and ball coordinate system ($\tilde{x}$, $\tilde{y}$, $\tilde{z}$). Both are orthogonal coordinate systems.}
    \label{img:coordinatesystems}
    \vspace{-0.2cm}
\end{figure}
\noindent Our model predicts two quantities: The 3D locations of the ball $\vec{r}(t_i)$ and the initial spin $\vec{\omega}(t_0)$.
To effectively describe these variables, we first define coordinate systems that facilitate both computation and human interpretation.
We introduce two coordinate systems: The \textit{world coordinate system} and the \textit{ball coordinate system}, both illustrated in Figure \ref{img:coordinatesystems}. \\[0.5ex]
\noindent The \textbf{world coordinate system}, remains fixed at the center of the table for each trajectory.
The $x$- and $y$-axes lie within the table plane, while the $z$-axis points upwards.
The unit vectors are defined as:
\begin{equation}
    \resizebox{0.9\linewidth}{!}{$
        \vec{e}_x = \begin{pmatrix}
            1 & 0 & 0
        \end{pmatrix}^\mathrm{T}
        \hspace{-0.15cm}
        \text{,}
        \hspace{0.3cm}
        \vec{e}_y = \begin{pmatrix}
            0 & 1 & 0
        \end{pmatrix}^\mathrm{T}
        \hspace{-0.15cm}
        \text{,} 
        \hspace{0.3cm}
        \vec{e}_z = \begin{pmatrix}
            0 & 0 & 1
        \end{pmatrix}^\mathrm{T}
        \hspace{-0.15cm}
        \text{.}
    $}
\end{equation}
This system is well-suited for describing and analyzing the ball's trajectory, which is why our model predicts the \textbf{3D trajectory} in this coordinate system.
However, while spin-related physical effects can be described in this system, interpreting individual spin components in an intuitive manner is difficult.
For this reason, we introduce a second coordinate system optimized for human interpretation. \\[0.5ex]
Unlike the world coordinate system, the \textbf{ball coordinate system} is adjusted once per trajectory.
This system is used to describe the ball's \textbf{initial spin} in a more interpretable way.
Its coordinate axes, denoted as $\tilde{x}$, $\tilde{y}$ and $\tilde{z}$, are defined as follows:
\begin{itemize}
    \item The origin is set at the ball's position in the first frame.
    \item The $\tilde{x}$-axis aligns with the ball's initial movement direction in the $x$-$y$ plane.
    \item The $\tilde{z}$-axis remains parallel to the $z$-axis, pointing upwards.
    \item The $\tilde{y}$-axis is orthogonal to both and lies in the $x$-$y$ plane.
\end{itemize}
Mathematical, the unit vectors are:
\begin{align}
    \label{eq:localcoords}
    \begin{split}
        &\vec{\Delta} = \begin{pmatrix}
            x(t_1) - x(t_0), &
            y(t_1) - y(t_0), &
            0
        \end{pmatrix}^\mathrm{T}
        \text{,}
        \\ \vspace{0.5cm}
        &\vec{e}_{\tilde{x}} = \frac{1}{\left| \vec{\Delta} \right|}\vec{\Delta} 
        \text{,}
        \hspace{0.5cm}
        \vec{e}_{\tilde{z}} = \vec{e}_z
        \text{,}
        \hspace{0.5cm}
        \vec{e}_{\tilde{y}} = \vec{e}_{\tilde{z}} \times \vec{e}_{\tilde{x}}
    \end{split}
\end{align}
with $t_i$ denoting the $i$-th timestep and $| \vec{\cdot} |$ being the euclidean norm of a vector. 
While the coordinate system is static during a trajectory, its center and orientation is different for each trajectory. \\[0.5ex]
If the trajectory is known, we can easily transform from one coordinate system to the other. 
For training we use the ground truth 3D trajectory for the transformation and during inference we rely on the 3D trajectory predictions. \\[0.5ex]
The ball coordinate system is chosen such that the main direction of the ball's movement is aligned with the $\tilde{x}$ axis. 
While there is also some movement in the $\tilde{z}$ direction, there is nearly no movement in the $\tilde{y}$ direction.
By adjusting the coordinate axes to the direction of the ball's initial movement, we ensure that the ball can be interpreted in a meaningful way.
Especially the rotation around the $\tilde{y}$ axis, denoted as $\omega_{\tilde{y}}$, is of high interest as it describes the top- or backspin of the ball.
In the next section, we will discuss the influence of the individual spin components in the ball coordinate system on the ball's trajectory in more detail.

\subsection{Observability of Spin Components}
Our model estimates the ball's initial spin solely from the observed 2D trajectory.
Therefore, if a spin component significantly affects the trajectory, we expect the model to learn it effectively.
The spin is characterized as a 3D vector $\vec{\omega}$ with its components describing the rotation around the corresponding axis of the ball coordinate system.
It influences the trajectory in two primary ways:
\begin{itemize}
    \item The \textbf{Magnus effect}: This force acts on a spinning ball with velocity $\vec{v}$, given by
    \begin{equation}
        \label{eq:Magnus}
        \vec{F} = k_\mathrm{M} \cdot \vec{\omega} \times \vec{v}
    \end{equation}
    where $k_\mathrm{M}$ is a constant that depends on the ball's surface properties.
    This force causes a deviation of the ball's trajectory perpendicular to $\vec{v}$ and $\vec{\omega}$ and affects the ball during the entire flight.
    \item \textbf{Friction at Bounce}: When the ball contacts the table, spin-induced friction generates force:
    \begin{equation}
        \vec{F} = k_\mathrm{F} \cdot \omega_{\tilde{x}} \vec{e}_{\tilde{y}} + k_\mathrm{F} \cdot \omega_{\tilde{y}} \vec{e}_{\tilde{x}}
    \end{equation}
    where $k_\mathrm{F}$ is a constant describing the ball's and table's surface properties, and $\vec{e}_{\tilde{x}}$ and $\vec{e}_{\tilde{y}}$ are the unit vectors of the ball coordinate system.
    This force only appears directly at the bounce and is not present during the flight.
\end{itemize}
The Magnus effect has the most significant influence on the trajectory, as it affects the ball during the entire flight.
As the main movement of the ball is in $\tilde{x}$ direction, only the orthogonal spin components $\omega_{\tilde{y}}$ and $\omega_{\tilde{z}}$ significantly influence the trajectory due to the cross product in Equation \ref{eq:Magnus}. \\[0.5ex]
The friction force only appears during the bounce and, thus, has a smaller influence on the trajectory.
It is governed by $\omega_{\tilde{x}}$ and $\omega_{\tilde{y}}$. \\[0.5ex]
As a result, $\omega_{\tilde{y}}$ and $\omega_{\tilde{z}}$ strongly impact the trajectory, while $\omega_{\tilde{x}}$ has a minor effect.
Therefore, we expect our model to learn the spin components $\omega_{\tilde{y}}$ and $\omega_{\tilde{z}}$ effectively, while the prediction of $\omega_{\tilde{x}}$ will be more challenging. \\[0.5ex]
The spin around the $\tilde{y}$ axis, denoted as $\omega_{\tilde{y}}$, describes the top- or backspin of the ball.
This component is easily interpretable and, therefore, of great importance in our analyses. \\[0.5ex]
In Figure \ref{img:spincomponents} we illustrate the effect of the spin components on the ball's trajectory in the image.
We simulate the trajectory of the ball, setting the same initial position and velocity for each trajectory, and only varying a single spin component in each case.
For each trajectory, we set one spin component to $-100 \, \text{Hz}$.
While the spin components $\omega_{\tilde{y}}$ and $\omega_{\tilde{z}}$ have a significant influence on the trajectory, $\omega_{\tilde{x}}$ has a negligible influence on the trajectory, which supports our previous discussion.

\section{Method}
\label{sec:method}
We train a neural network to predict the ball's 3D trajectory and initial spin from its 2D trajectory. 
Each trajectory usually goes from the table tennis shot over a single touchdown on the tennis table to the next touch.
The general pipeline of our method is shown in Figure \ref{fig:pipeline}.

\subsection{Training Objective}
Our neural network is trained using a large dataset of simulated synthetic data, enabling \textbf{direct supervision with synthetic ground truth}. The loss function for predicting the 3D trajectory is defined as:
\begin{equation}
    \mathcal{L}_\text{trajectory} = \frac{1}{T} \sum\limits_{i = 0}^{T-1} \left\| \vec{r}_\text{pred}(t_i) - \vec{r}_\text{gt}(t_i) \right\|^2
\end{equation}
where $\vec{r}_\text{pred}(t_i)$ represents the predicted 3D position of the ball at time $t_i$, and $\vec{r}_\text{gt}(t_i)$ denotes the corresponding ground truth position.
The loss is averaged over all time steps $T$.
To evaluate the accuracy of the predicted initial spin, we define the spin loss as:
\begin{equation}
    \mathcal{L}_\text{spin} = \left\| \vec{\omega}_\text{pred} - \vec{\omega}_\text{gt} \right\|^2
\end{equation}
where $\vec{\omega}_\text{pred}$ and $\vec{\omega}_\text{gt}$ are the predicted and ground truth initial spins, respectively.
Consequently, the total loss is given by
\begin{equation}
    \label{eq:loss}
    \mathcal{L} = \mathcal{L}_\text{trajectory} + \mathcal{L}_\text{spin} \, .
\end{equation}
This formulation ensures that both the trajectory prediction and initial spin estimation are optimized simultaneously.
Because both losses are in the same order of magnitude, we do not introduce additional weighting factors.

\subsection{Base Architecture}
\label{sec:architecture}
Our proposed architecture consists of an \textit{embedding module} and a \textit{Spin Prediction Transformer (SPT)}. 
Given a sequence of length $T$, the model takes as input the set of the 2D ball positions and 13 table keypoints at each time $t_i$. \\[0.5ex]
The embedding module transforms this 2D information into a $d$-dimensional \textit{location token} $l_i \in \mathbb{R}^d$ for each $t_i$.
Additionally, a learnable \textit{spin token} $s \in \mathbb{R}^d$ is prepended, resulting in the sequence $\{s, l_0, ..., l_{T-1}\} \in \mathbb{R}^{(T+1)\times  d}$ that is then processed by the SPT. \\[0.5ex]
After the final transformer layer, a \textit{position head} is applied to each transformed location token individually, predicting the 3D positions $\{\vec{r}(t_0), ..., \vec{r}(t_{T-1})\} \in \mathbb{R}^{T \times 3}$.
Similarly, a \textit{spin head} is applied to $s$, predicting the initial spin $\vec{\omega} \in \mathbb{R}^3$.
The overall architecture is illustrated in Figure \ref{fig:pipeline}.

\vspace{-0.3cm}
\paragraph{Token Embeddings} 
We explore three embedding strategies for computing the location tokens $l_i$:
\begin{itemize}
    \item \textbf{Concatenation Method:} Concatenates the 2D coordinates of all 14 points (ball position + 13 table points) into a single vector, followed by an MLP with one hidden layer to obtain the location token $l_i$.
    \item \textbf{Dynamic Method:} Treats each of the 14 points as a separate token, projects them into a $d$-dimensional space using an MLP with one hidden layer, and condenses the information via a 4-layer transformer encoder. The transformed ball position token is used as $l_i$, discarding the other tokens.
    \item \textbf{Context-Free Method:} Uses only the 2D ball position as input. An MLP with one hidden layer projects the ball position into the $d$-dimensional space to obtain the location token $l_i$. This method serves as a baseline to check if the information from the table keypoints is needed.
\end{itemize}
These methods are visualized in Figure \ref{fig:token_embedding} of the supplementary material.

\vspace{-0.3cm}
\paragraph{Spin Prediction Transformer}
\begin{figure}[t]
    \centering
    \begin{subfigure}{\linewidth}
        \centering
        \includegraphics[width=\linewidth]{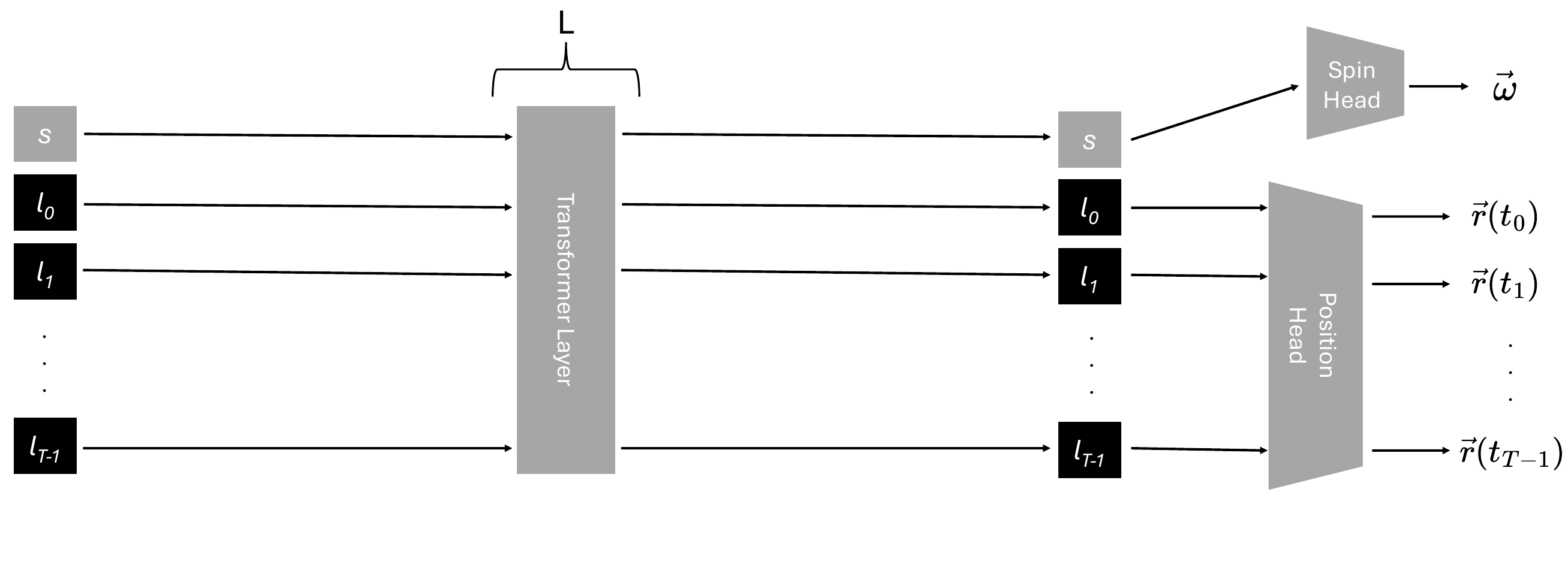}
        \vspace{-0.5cm}
        \caption{Single-Stage Model}
        \label{fig:singlestage}
    \end{subfigure}
    \begin{subfigure}{\linewidth}
        \centering
        \includegraphics[width=\linewidth]{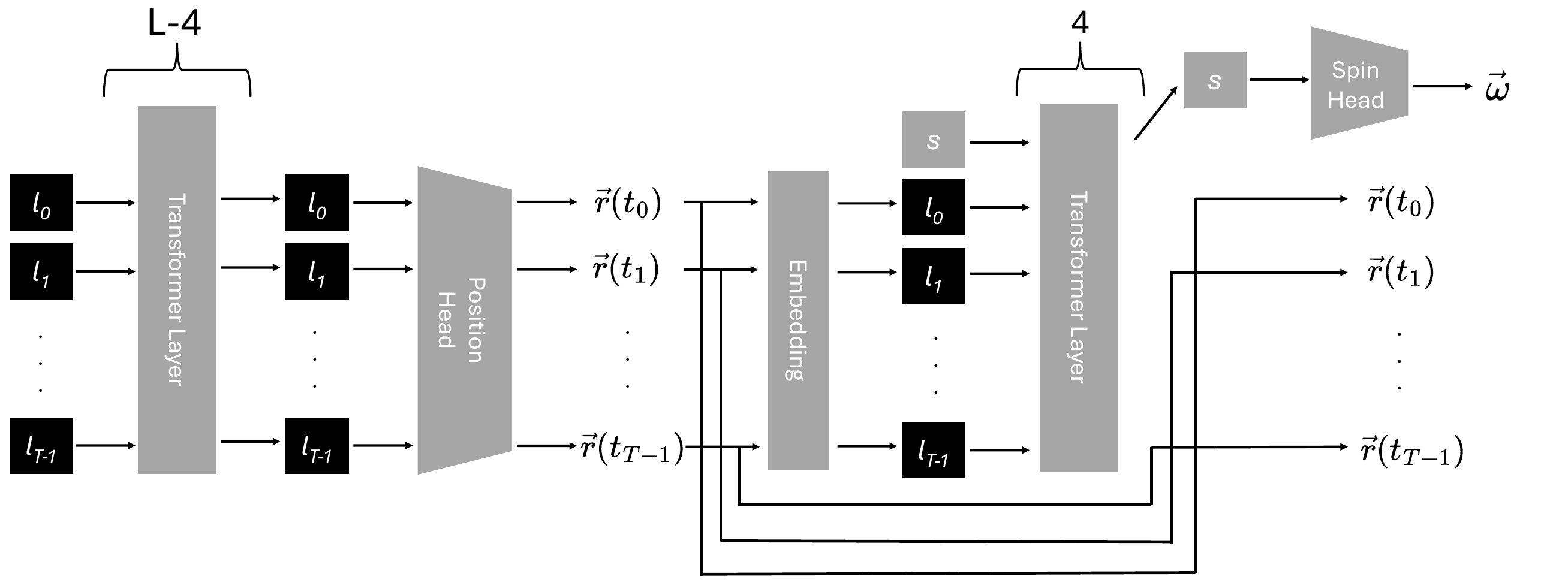}
        \vspace{-0.3cm}
        \caption{Two-Stage Model}
        \label{fig:multistage}
    \end{subfigure}
    \begin{subfigure}{\linewidth}
        \centering
        \includegraphics[width=\linewidth]{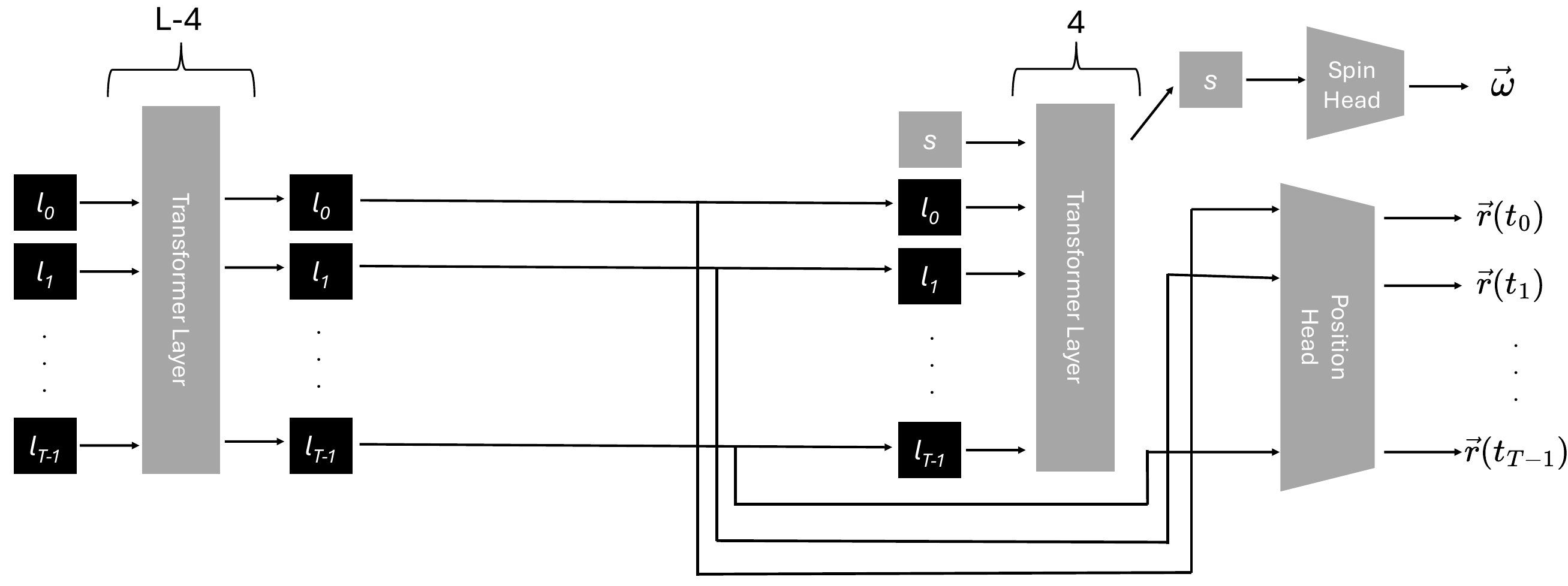}
        \vspace{-0.3cm}
        \caption{Connect-Stage Model}
        \label{fig:connectstage}
    \end{subfigure}
    \vspace{-0.2cm}
    \caption{Illustration of the 3 different SPT architectures. In the single-stage model, the trajectory is predicted jointly with the spin. 
    The two-stage model predicts only the 3D positions $\vec{r}(t_i) \in \mathcal{R}^3$ in the first stage and uses these predictions to estimate the spin in the second stage.
    The connect-stage model uses the transformed tokens $l_i \in \mathcal{R}^d$ as input for the second stage.
    }
    \label{fig:rpt_architecture}
    \vspace{-0.2cm}
\end{figure}
The SPT is an encoder-only transformer with $L$ layers. 
It processes the sequence $\{s, l_0, ..., l_{T-1}\}$ and outputs $T+1$ tokens.
A position head is applied to each transformed location token $l_i$ for predicting the 3D positions $\vec{r}(t_i) \in \mathbb{R}^3$, while the spin head processes the spin token $s$ to predict the initial spin $\vec{\omega} \in \mathbb{R}^3$. \\[0.5ex]
We explore three SPT architectures (see Figure \ref{fig:rpt_architecture}):
\begin{itemize}
    \item \textbf{Single-Stage Model:} Prepends the spin token to the location tokens before the first transformer layer. It predicts positions and spin jointly by processing all tokens together. However, this approach does not explicitly enforce a dependency between trajectory and spin.
    \item \textbf{Two-Stage Model:} Enforces a physical bottleneck by first predicting 3D positions $\vec{r}(t_i) \in \mathbb{R}^3$ with $L - 4$ transformer layers (no spin token is prepended), then using these positions as the only input for spin prediction in a second stage (4-layer transformer). Since spin and trajectory are directly linked, spin alters the trajectory and trajectory encodes spin information, our model explicitly captures this dependency. While this architecture mimics this physical connection, any noise in the predicted positions directly affects spin accuracy.
    \item \textbf{Connect-Stage Model:} Addresses the rigid bottleneck in the two-stage model by using the transformed tokens $l_i \in \mathbb{R}^d$ before the application of the position head as input of the second stage. This softens the spin's dependency on the trajectory prediction, while still maintaining the physically motivated structure.
\end{itemize}

\vspace{-0.3cm}
\paragraph{Further implementation details}
While the positions are always predicted in the world coordinate system, we can predict the initial spin either in the world or the ball coordinate system.
If predicted in the world coordinate system, it is converted to the ball coordinate system, which is described by Equation \ref{eq:localcoords}. \\[0.5ex]
We use Rotary Positional Embeddings (RoPE) \cite{RoPE}, which have been recently gaining popularity in the field of language processing \cite{llama3,deepseekvl,phi1}, to encode the order of the tokens.
Instead of adding absolute positional embeddings \cite{transformer}, a rotation is applied to the queries and keys before each attention computation.
The strength of the rotation is dependent on the position in the sequence, thus the computed attention scores include positional information. \\[0.5ex]
We implement and test different hyperparameters of the architecture and define various sizes of the model as shown in Table \ref{tab:sizes} of the supplementary material.
We use a model with $L=16$ and $d=128$, referred to as \textit{large} model, for all experiments.

\section{Data}
\label{sec:data}

\subsection{Synthetic Data Generation}
We generate physically accurate table tennis trajectories using MuJoCo \cite{mujoco}, with a particular focus on realistic ball bounces, which significantly impact data quality. 
Following \cite{AchievingHumanLevel}, which trained a human-level robot using synthetic data alone, we adopt the same simulation parameters. \\[0.5ex]
Trajectories are generated by sampling initial position, velocity, and spin, followed by simulating the trajectory.
Only valid trajectories are kept. 
A valid trajectory starts on the left side, bounces once on the opponent's side, and ends on the right, ensuring full visibility in the camera frame.
We collect \textbf{50,000} valid trajectories, split into \textbf{70\%} training, \textbf{10\%} validation, and \textbf{20\%} test set. 
To ensure robustness, we vary camera parameters during training while using real-data camera parameters for validation and testing.
We illustrate some sampled camera parameters in Figure \ref{fig:sampled_cameras} of the supplementary material. \\[0.5ex]
As ground-truth 3D trajectories and rotations are available in synthetic data, we compute the \textit{spin error} as
\begin{equation}
    \Delta \vec{\omega} = \frac{1}{N}\sum\limits_{j=1}^{N} || \vec{\omega}_{\text{pred},j} - \vec{\omega}_{\text{gt},j} ||_2
\end{equation}
and the \textit{3D trajectory error} as
\begin{equation}
    \Delta \vec{r}_\text{world} = \frac{1}{N}\sum\limits_{j=1}^{N} \frac{1}{T_j} \sum\limits_{i = 0}^{T_j - 1} || \vec{r}_{\text{pred},j}(t_i) - \vec{r}_{\text{gt},j}(t_i) ||_2
\end{equation}
where $N$ is the number of trajectories, $j$ indexes the different trajectories, $t_i$ indexes time steps of a trajectory, $T_j$ is the length of trajectory $j$, and $|| \cdot ||_2$ is the Euclidean vector norm.

\subsection{Real Data}
Although training is solely conducted on synthetic data, we assess generalization on real table tennis broadcasts. We manually annotate \textbf{50} 2D ball trajectories from six WTTF matches, totaling \textbf{1197} annotated frames. Table keypoints are annotated once per trajectory and reused across frames. The videos have a $2560 \times 1440$ resolution at \SI{50}{Hz}. \\[0.5ex]
While obtaining 3D ground truth for real data is infeasible, we manually label spin direction (top-/backspin) by analyzing paddle positions in the image (Figure \ref{fig:spinannotation}, supplementary material). We evaluate classification accuracy as
\begin{equation} 
    \textit{acc} = \frac{1}{N} \sum\limits_{j = 1}^N \delta_{c_j, \text{sign}( \omega_{\tilde{y}, j} )} 
\end{equation}
where $c_j$ is the annotated class (1 for topspin, -1 for backspin), $\omega_{\tilde{y}, j}$ is the predicted spin, $\text{sign}(\cdot )$ is the sign function, and $\delta_{a, b}$ is 1 if $a=b$, otherwise $0$.
Similarly, we also compute the macro F1-score. \\[0.5ex]
To convert the estimated $\omega_{\tilde{y}}$ into a binary classification, we set a fixed threshold at $\omega_{\tilde{y}}=0$, which is physically justified since the sign of $\omega_{\tilde{y}}$ determines the spin direction.
However, this strict threshold may overly penalize trajectories with weak spin, where small prediction errors lead to entirely incorrect classifications.
In standard binary classification tasks, the ROC-AUC metric \cite{rocauc} avoids reliance on a fixed threshold, making it a more robust performance measure.
While accuracy and F1-score remain our primary metrics due to the known threshold, we also evaluate ROC-AUC, as it provides a more nuanced assessment, especially for predictions near zero.
As we will see in the experiments, the key threshold values in the ROC-AUC computation are indeed close to zero, further supporting its relevance (see Figure \ref{fig:roc}) \\[0.5ex]
For trajectory evaluation, we project predicted 3D paths into 2D image coordinates and compute the \textit{2D reprojection error} $\Delta \vec{r}_\text{img}$:
\begin{equation}
    \resizebox{0.88\linewidth}{!}{$
    \Delta \vec{r}_\text{img} = \frac{1}{D} \frac{1}{N}\sum\limits_{j=1}^{N} \frac{1}{T_j} \sum\limits_{i = 0}^{T_j - 1} || \mathcal{P} \left( \vec{r}_{\text{pred},j}(t_i) \right) - \vec{r}_{\text{gt2D},j}(t_i) ||_2 
    $}
\end{equation}
where $\mathcal{P} \in \mathcal{R}^{2\times3}$ projects 3D coordinates into 2D image positions. The matrix $\mathcal{P}$ is estimated using the table point annotations (see Section \ref{sec:regresscameramatrices} in the supplementary material).
Instead of evaluating the absolute pixel error, we divide through the image diagonal length $D = \frac{1}{\sqrt{H^2+W^2}}$ (with $H \times W$ being the video resolution). 
Consequently, we obtain a relative error that is not dependent on the video resolution and, thus, allows for better human interpretability. \\[0.5ex]
Despite lacking direct 3D annotations, our methodology enables indirect assessment of the generalization ability through classification and projected trajectory comparisons.

\subsection{Data Augmentations}
\label{sec:augmentations}
By using a smart data representation instead of raw visual data as input and utilizing physically correct synthetic data, we enable the model to generalize to real broadcast.
However, three key challenges still introduce inaccuracies:
\begin{itemize}
    \item \textbf{Motion Blur}: The ball exhibits strong motion blur, making precise localization difficult and introducing errors absent in synthetic data.
    \item \textbf{Sudden Trajectory Stops}: In real matches, the opponent's hit often terminates the trajectory suddenly, unlike in synthetic simulations where we do not simulate the opposing player.
    \item \textbf{Annotation Errors}: Manually labeled table points may contain slight inaccuracies, whereas synthetic training data is always precise.
\end{itemize}
To mitigate these effects, we introduce three augmentations: \textit{motion blur}, \textit{sudden trajectory stop}, and \textit{gaussian blur}.
The implementation of the different augmentations is described in more detail in Section \ref{sec:augmentationdetails} of the supplementary material.

\section{Results}
\label{sec:results}
Each model is implemented in PyTorch \cite{pytorch} and trained on a single NVIDIA H100 GPU.
We use a fixed learning rate of $10^{-4}$ and a batch size of 64.
Model weights are optimized with ADAM \cite{ADAM}, and an Exponential Moving Average (EMA) with a decay of $0.999$ \cite{EMA} is applied.
Training runs for 800 epochs, and we select the model with the best $\Delta \vec{\omega}$ score on the synthetic validation set. \\[0.5ex]
We consistently use the large model with rotary positional encodings, predicting spin in the world coordinate system.
Additional ablations on these parameters are provided in the supplementary material. \\[0.5ex]
Our primary focus is on real-data performance, as generalization to real-world scenarios is crucial.
Synthetic performance is not relevant for practical applications, but is included in all tables.
We further discuss the significance of synthetic results in Section \ref{sec:evalarchitecture}.

\subsection{Evaluation of SPT Architectures}
\label{sec:evalarchitecture}
We compare the SPT architectures introduced in Section \ref{sec:architecture} without data augmentations, using the concatenation method for token embeddings.
Results are shown in Table \ref{tab:architecture}. \\[0.5ex]
\begin{table}[h]
    \centering
    \resizebox{\linewidth}{!}{ 
    \begin{tabular}{l|c|c|c|c|c|c}
        \toprule
         & \multicolumn{2}{c|}{Synthetic} & \multicolumn{4}{c}{Real} \\
        Method & $\Delta \vec{\omega}$ $\downarrow$ & $\Delta \vec{r}_\text{world}$ $\downarrow$ & $\textit{acc}$ $\uparrow$ & $\mathrm{F}_1$ $\uparrow$ & ROC-AUC $\uparrow$ & $\Delta \vec{r}_\text{img}$ $\downarrow$ \\
        \midrule
        single-stage & \SI{11.7}{Hz} & \SI{35.2}{cm} & \SI{58.0}{\%} & \SI{0.440}{} & \SI{0.669}{} & \SI{7.67}{\%} \\ 
        two-stage & \SI{56.4}{Hz} & \SI{4.6}{cm} & \bfqty{80.0}{\%} & \bfqty{0.799}{} & \SI{0.807}{} & \SI{0.72}{\%} \\ 
        connect-stage & \SI{31.0}{Hz} & \SI{3.6}{cm} & \SI{74.0}{\%} & \SI{0.740}{} & \bfqty{0.838}{} & \bfqty{0.53}{\%} \\ 
        \bottomrule
    \end{tabular}
    }
    \caption{Comparison of different SPT architectures. The best results on the real data are highlighted in bold.}
    \label{tab:architecture}
\end{table}
The single-stage model performs well for spin prediction on synthetic data but fails to generalize to real data.
Its 3D trajectory predictions are also subpar.
The two-stage model, in contrast, achieves strong results on real data despite weaker synthetic performance.
The connect-stage model slightly lags behind the two-stage model in accuracy and macro F1 score but outperforms in ROC-AUC and 2D reprojection error ($\Delta \vec{r}_\text{img}$). \\[0.5ex]
Both two-stage and connect-stage models generalize well.
The two-stage model is preferable for spin classification, while the connect-stage model excels in 3D trajectory prediction.
Conclusively, both models are suitable for our task.
As the connect-stage model subjectively offers the best balance, we use it for further experiments. \\[0.5ex]
Synthetic results do not reliably predict real-world performance.
For instance, the single-stage model excels on synthetic data but fails on real data, whereas the two-stage model shows the opposite trend.
This suggests that the physically motivated bottleneck in the two-stage and connect-stage models improves generalization.
By constraining the model architecture, we enhance real-data applicability despite training solely on synthetic data.
Hence, subsequent experiments focus exclusively on real-data performance.

\subsection{Evaluation of Token Embedding Methods}
We assess different token embedding methods discussed in Section \ref{sec:architecture} using the connect-stage architecture without data augmentations.
Results are presented in Table \ref{tab:embeddings}. \\[0.5ex]
\begin{table}[h]
    \centering
    \resizebox{\linewidth}{!}{ 
    \begin{tabular}{l|c|c|c|c|c|c}
        \toprule
         & \multicolumn{2}{c|}{Synthetic} & \multicolumn{4}{c}{Real} \\
        Method & $\Delta \vec{\omega}$ $\downarrow$ & $\Delta \vec{r}_\text{world}$ $\downarrow$ & $\textit{acc}$ $\uparrow$ & $\mathrm{F}_1$ $\uparrow$ & ROC-AUC $\uparrow$ & $\Delta \vec{r}_\text{img}$ $\downarrow$ \\
        \midrule
        context free & \SI{41.5}{Hz} & \SI{54.9}{cm} & \SI{66.0}{\%} & \SI{0.649}{} & \SI{0.670}{} & \SI{4.37}{\%} \\ 
        dynamic & \SI{27.7}{Hz} & \SI{3.3}{cm} & \bfqty{84.0}{\%} & \bfqty{0.836}{} & \bfqty{0.911}{} & \SI{0.56}{\%} \\ 
        concatenation & \SI{31.0}{Hz} & \SI{3.6}{cm} & \SI{74.0}{\%} & \SI{0.740}{} & \SI{0.838}{} & \bfqty{0.53}{\%} \\ 
        \bottomrule
    \end{tabular}
    }
    \caption{Comparison of different embedding methods. The best results on the real data are highlighted in bold.}
    \label{tab:embeddings}
\end{table}
The context-free method performs the worst across all metrics.
Without table keypoints, it fails to capture sufficient scene context, confirming that the 2D trajectory alone is inadequate for an end-to-end method.
In contrast, both dynamic and concatenation methods yield strong results, demonstrating the importance of table keypoints as input. \\[0.5ex]
The dynamic method is best for spin classification, while the concatenation method slightly outperforms in 3D trajectory prediction.
We attribute the dynamic method's success to its more complex embedding process, which leverages transformer layers to integrate keypoint and ball position data.
However, this increases the parameter count, computational cost and results in sensitivity to hyperparameters as well as less stable training.
Given its robustness and efficiency, we select the concatenation method for our further experiments. \\[0.5ex]

\subsection{Evaluation of Data Augmentations}
We evaluate the three data augmentation techniques described in Section \ref{sec:augmentations} using the connect-stage architecture with concatenation embeddings.
Results are in Table \ref{tab:augmentations}. \\[0.5ex]
\begin{table}[h]
    \centering
    \resizebox{\linewidth}{!}{ 
    \begin{tabular}{c|c|c|c|c|c|c|c|c}
        \toprule
        \multicolumn{3}{c|}{Method} & \multicolumn{2}{c|}{Synthetic} & \multicolumn{4}{c}{Real} \\
        \makecell{motion\\blur} & \makecell{sudden\\end} & \makecell{gaus.\\blur} & $\Delta \vec{\omega}$ $\downarrow$ & $\Delta \vec{r}_\text{world}$ $\downarrow$ & $\textit{acc}$ $\uparrow$ & $\mathrm{F}_1$ $\uparrow$ & ROC-AUC $\uparrow$ & $\Delta \vec{r}_\text{img}$ $\downarrow$ \\
        \midrule
        \ding{53} & \ding{53} & \ding{53} & \SI{31.0}{Hz} & \SI{3.6}{cm} & \SI{74.0}{\%} & \SI{0.740}{} & \SI{0.838}{} & \SI{0.53}{\%} \\
        \ding{51} & \ding{53} & \ding{53} & \SI{44.5}{Hz} & \SI{5.0}{cm} & \SI{88.0}{\%} & \SI{0.875}{} & \SI{0.969}{} & \SI{0.55}{\%} \\
        \ding{53} & \ding{51} & \ding{53} & \SI{31.7}{Hz} & \SI{3.8}{cm} & \bfqty{96.0}{\%} & \bfqty{0.959}{} & \bfqty{0.990}{} & \SI{0.34}{\%} \\
        \ding{53} & \ding{53} & \ding{51} & \SI{42.5}{Hz} & \SI{4.2}{cm} & \SI{80.0}{\%} & \SI{0.800}{} & \SI{0.898}{} & \SI{0.64}{\%} \\

        \ding{51} & \ding{51} & \ding{51} & \SI{48.7}{Hz} & \SI{5.5}{cm} & \SI{92.0}{\%} & \SI{0.917}{} & \bfqty{0.990}{} & \bfqty{0.19}{\%} \\
        \bottomrule
    \end{tabular}
    }
    \caption{Comparison of different data augmentations. 
    \ding{51} indicates the application of the augmentation, while \ding{53} indicates the absence of the augmentation.
    The best results on the real data are highlighted in bold.}
    \label{tab:augmentations}
    \vspace{-0.2cm}
\end{table}
All augmentations improve generalization to real data, with sudden end augmentation providing the most substantial gains across all metrics.
Motion blur also enhances performance significantly, while Gaussian blur offers only slight improvements.
This demonstrates that making synthetic data more realistic significantly boosts real-world performance. \\[0.5ex]
Since each augmentation contributes positively, we combine them in the last row of Table \ref{tab:augmentations}.
This results in a notable reduction in 2D reprojection error while maintaining strong spin prediction accuracy.
This model gives a good trade-off between spin classification and 3D trajectory prediction and, thus, this combination is referred to as the \textit{best model} in the following experiments. \\[0.5ex]
Although the model already performs well on real data without any augmentations, applying them further enhances performance.
Our best model nearly achieves perfect results, indicating that augmentations effectively bridge the gap between synthetic and real data.
They are the final step to enable generalization from synthetic training alone.

\subsection{Detailed Evaluation of Best Model}
\label{sec:detailedeval}
In this section, we analyze the performance of our best model in greater detail.
This model, corresponding to the last row in Table \ref{tab:augmentations}, employs the connect-stage architecture with concatenation token embeddings and incorporates all data augmentations during training.
Our focus is on evaluating its effectiveness in spin classification and 3D trajectory prediction on real data. 
\begin{figure}[ht]
    \vspace{-0.5cm}
    \centering
    \begin{subfigure}[t]{0.37\linewidth}
        \centering
        \includegraphics[width=\linewidth]{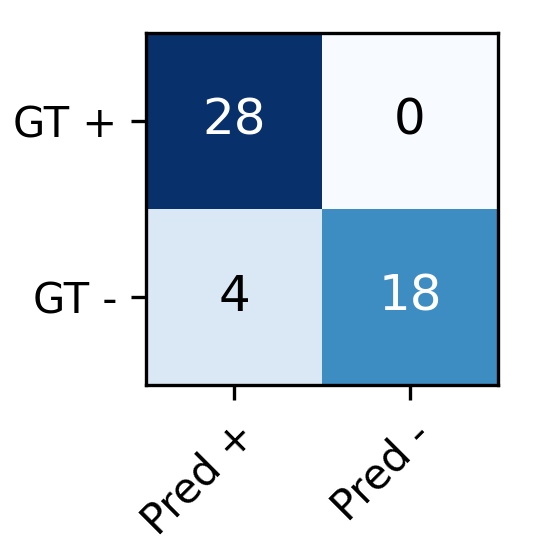}
        \caption{Confusion matrix for spin classification on real data. Topspin is \textbf{+} and backspin \textbf{-}.}
        \label{fig:confusion}
    \end{subfigure}
    \hfill
    \begin{subfigure}[t]{0.59\linewidth}
        \centering
    \includegraphics[width=\linewidth]{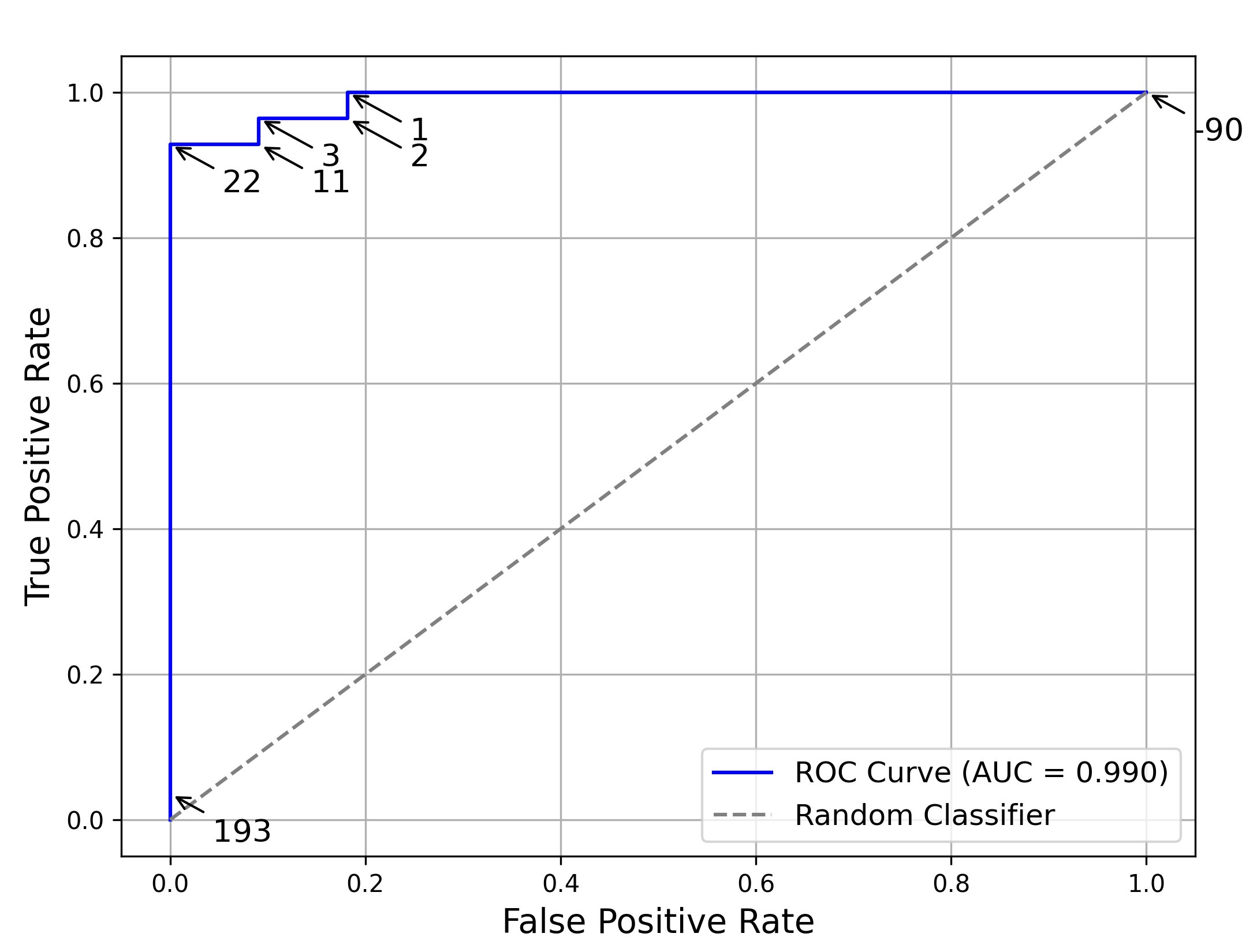}
    \caption{ROC curve for spin classification on real data.
    The arrows indicate the threshold values in Hz for the predicted $\omega_{\tilde{y}}$.}
    \label{fig:roc}
    \end{subfigure}
    \caption{Confusion matrix and ROC plot for the best model on the real dataset.}
    \label{fig:confusionroc}
\end{figure}
\vspace{-0.6cm}
\paragraph{Spin Classification Performance}
Figure \ref{fig:confusionroc} presents the confusion matrix and the ROC curve for spin classification on the real test dataset.
The confusion matrix reveals that the model distinguishes well between topspin and backspin, though a slight bias towards topspin is noticeable.
This is also reflected in the ROC curve, where the threshold values for predicted $\omega_{\tilde{y}}$ are subtly shifted towards topspin. \\[0.5ex]
Despite this bias, the threshold values remain close to the physically correct threshold of \SI{0}{Hz}, suggesting that the model's predictions are not significantly skewed.
Even though we cannot evaluate the exact values of $\omega_{\tilde{y}}$ on the real data, the smooth and consistent behavior observed in the ROC thresholds indicates that the model produces reasonable and stable predictions.
Overall, these results confirm the model's reliability in spin classification.

\vspace{-0.2cm}
\paragraph{3D Trajectory Prediction Performance}
\begin{figure}
    \centering
    \begin{subfigure}[t]{0.49\linewidth}
        \centering
        \includegraphics[width=\linewidth]{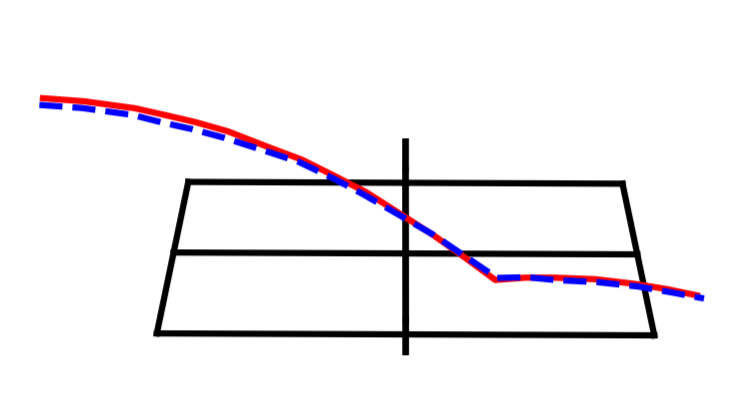}
    \end{subfigure}
    \hfill
    \begin{subfigure}[t]{0.49\linewidth}
        \centering
        \includegraphics[width=\linewidth]{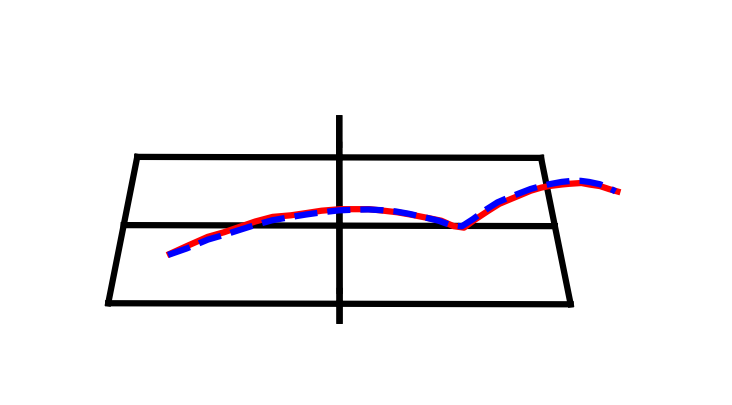}
    \end{subfigure}
    \caption{Comparison of reprojection of the predicted trajectory (dashed blue) with the annotated 2D trajectory (solid red). We present two examples on the real dataset.}
    \label{fig:trajectory}
    \vspace{-0.0cm}
\end{figure} ~
\nolinebreak In Figure \ref{fig:trajectory}, we illustrate $2$ predicted trajectories on the real dataset.
To compare a predicted trajectory with the corresponding annotated 2D trajectory, we calculate the reprojection of the predicted 3D trajectory onto the 2D image.
In both examples, the predicted trajectory closely follows the annotated ground truth, demonstrating strong alignment between the two.
This suggests that the model effectively captures the underlying 3D motion of the ball and generalizes well to real-world data. \\[0.5ex]
The high accuracy in trajectory prediction confirms that the model correctly generalizes to real data, showing its suitability for practical applications.

\section{Conclusion}
\label{sec:conclusion}
In this paper, we introduced a method for predicting the 3D trajectory and initial spin of a table tennis ball in broadcast videos.
A key aspect of our approach is that the model is trained exclusively on synthetic data, yet it generalizes remarkably well to real-world footage.
This strong generalization is achieved through a combination of a carefully designed data representation, physically grounded synthetic training data, and problem-specific data augmentations.
Notably, our results demonstrate that these straightforward, yet effective steps are sufficient to bridge the gap between synthetic and real data.
Our method enables detailed analysis of a player's technique using standard monocular RGB videos, making advanced performance evaluation more accessible.
It can be applied both by professionals analyzing broadcast footage and by amateurs using a simple smartphone camera, broadening its usability across different levels of expertise.

\newpage

{
    \small
    \bibliographystyle{ieeenat_fullname}
    \bibliography{main}
}

\appendix
\clearpage
\maketitlesupplementary

\section{Further Architecture Details}
\label{sec:furtherarchitecturedetails}
\begin{figure}[h]
    \centering
    \begin{subfigure}{\linewidth}
        \centering
        \includegraphics[width=\linewidth]{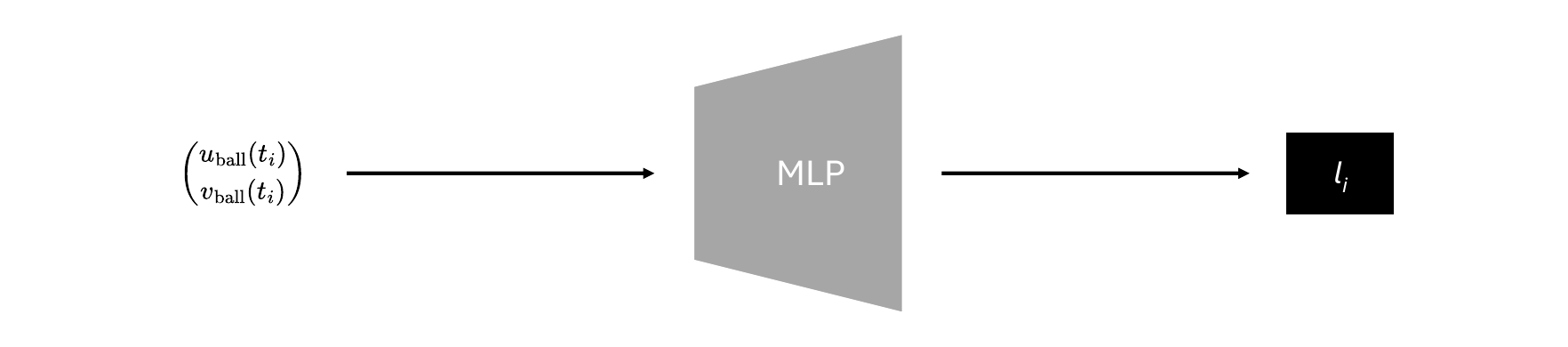}
        \caption{Context-Free Method}
        \label{fig:contextfreeembedding}
    \end{subfigure}
    \begin{subfigure}{\linewidth}
        \centering
        \includegraphics[width=\linewidth]{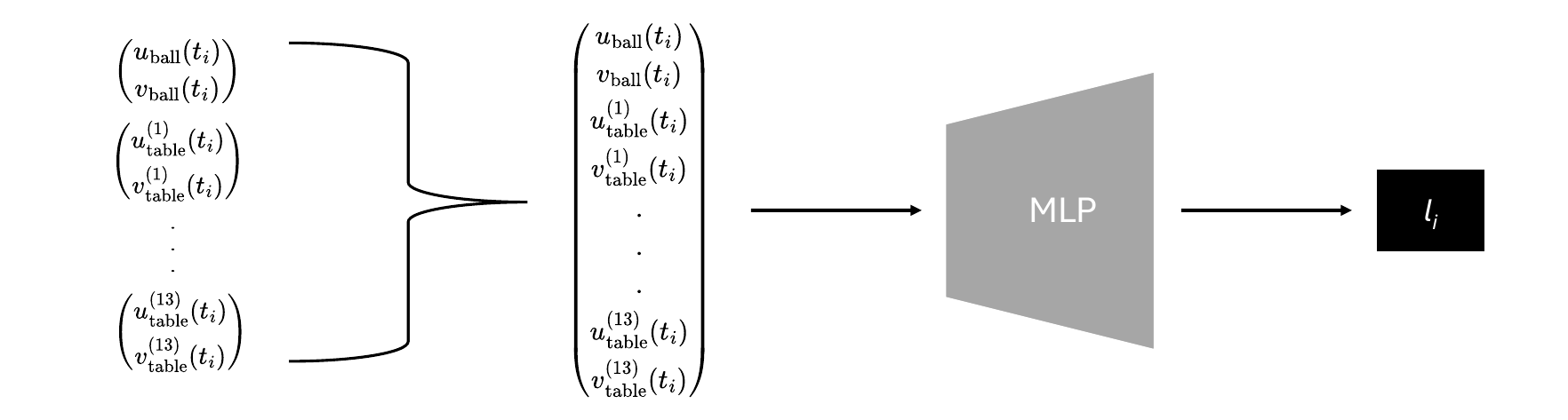}
        \caption{Concatenation Method}
        \label{fig:concatenationembedding}
    \end{subfigure}
    \begin{subfigure}{\linewidth}
        \centering
        \includegraphics[width=\linewidth]{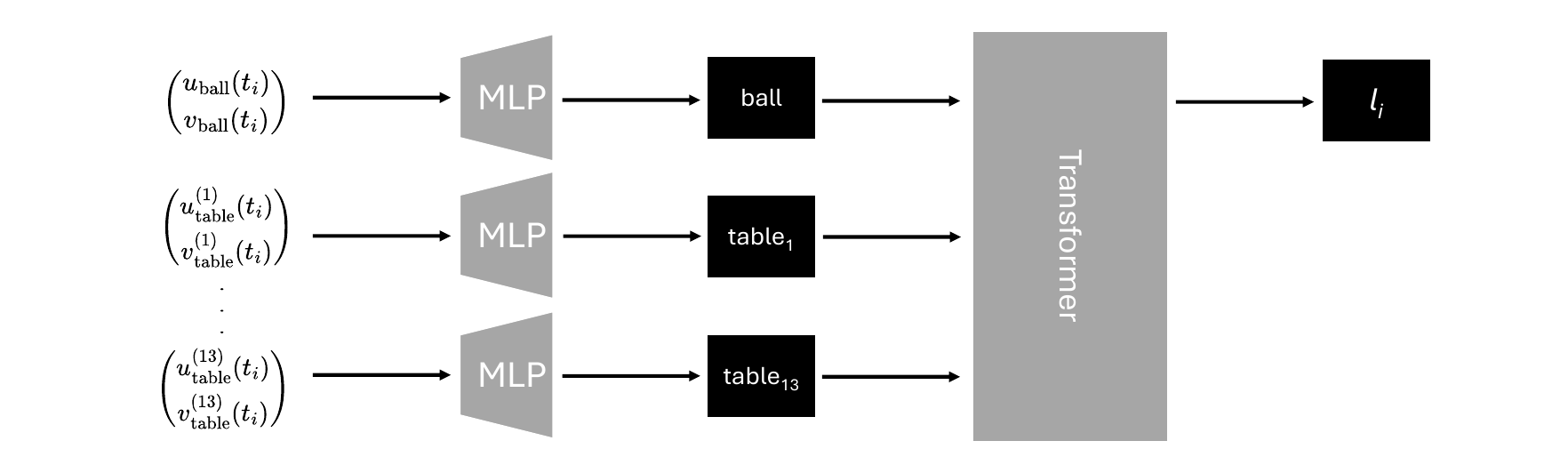}
        \caption{Dynamic Method}
        \label{fig:dynamicembedding}
    \end{subfigure}
    \caption{Token embedding methods. Input to the embedding layers are the 2D coordinates of the ball and the 13 table keypoints. The output is the location token $l_i$. The embedding layer is applied for each time step $t_i$ separately.}
    \label{fig:token_embedding}
\end{figure}
\noindent Figure \ref{fig:token_embedding} provides an overview of the different token embedding strategies utilized in our model. 
Each method processes the 2D coordinates of the ball along with 13 table keypoints to generate the location token $l_i$ at time $t_i$. 
The embedding operation is applied separately for each time step $t_i$.
\begin{itemize}
    \item \textbf{Context-Free Method:} The context-free method directly embeds the ball coordinates via a multilayer perceptron (MLP) without using any table keypoints.
    \item \textbf{Concatenation Method:} The 2D coordinates of all 14 points are concatenated into a single vector. This vector is then transformed into a location token via an MLP.
    \item \textbf{Dynamic Method:} Instead of direct concatenation, a small transformer encoder processes the table keypoints and condenses their information into the ball position token. This token is then used as location token $l_i$, the other tokens are discarded.
\end{itemize}
\noindent To evaluate model performance across different architectural complexities, we vary the number of transformer layers $L$, the number of attention heads $H$, and the embedding dimension $d$.
Table \ref{tab:sizes} summarizes the configurations explored.
\begin{table}[ht]
    \centering
    \resizebox{\linewidth}{!}{
    \begin{tabular}{c|c|c|c|c}
    \textbf{Size} & \textbf{Layers $L$} & \textbf{Heads $H$} & \textbf{Embedding Dimension $d$} & \textbf{Number of Parameters}  \\
    \hline
    Small & 8 & 4 & 32 & $0.06\times 10^6$ \\
    Base & 12 & 4 & 64 & $0.3\times 10^6$ \\
    Large & 16 & 4 & 128 & $1.6\times 10^6$ \\
    Huge & 16 & 8 & 192 & $3.2\times 10^6$ \\
    \end{tabular}
    }
    \caption{Transformer architecture variants. The number of trainable parameters is calculated for the model with connect-stage SPT architecture and concatenation token embedding module.}
    \label{tab:sizes}
    \end{table}

\section{Annotation Details}
\label{sec:annotationdetails}
\begin{figure}[h]
    \centering
    \includegraphics[width=0.9\linewidth]{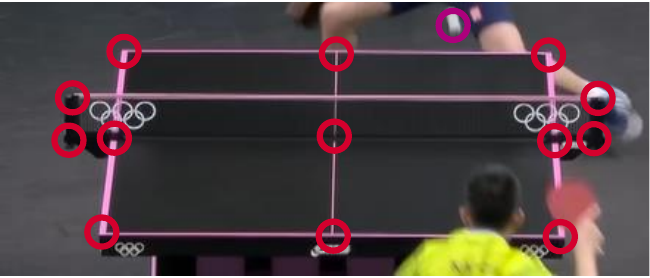}
    \caption{The 13 table keypoints (red circles) and the ball position (purple circle) are highlighted.}
    \label{img:tablekeypoints}
\end{figure}

\begin{figure}[h]
    \centering
    \begin{subfigure}[t]{0.9\linewidth}
        \includegraphics[width=\linewidth]{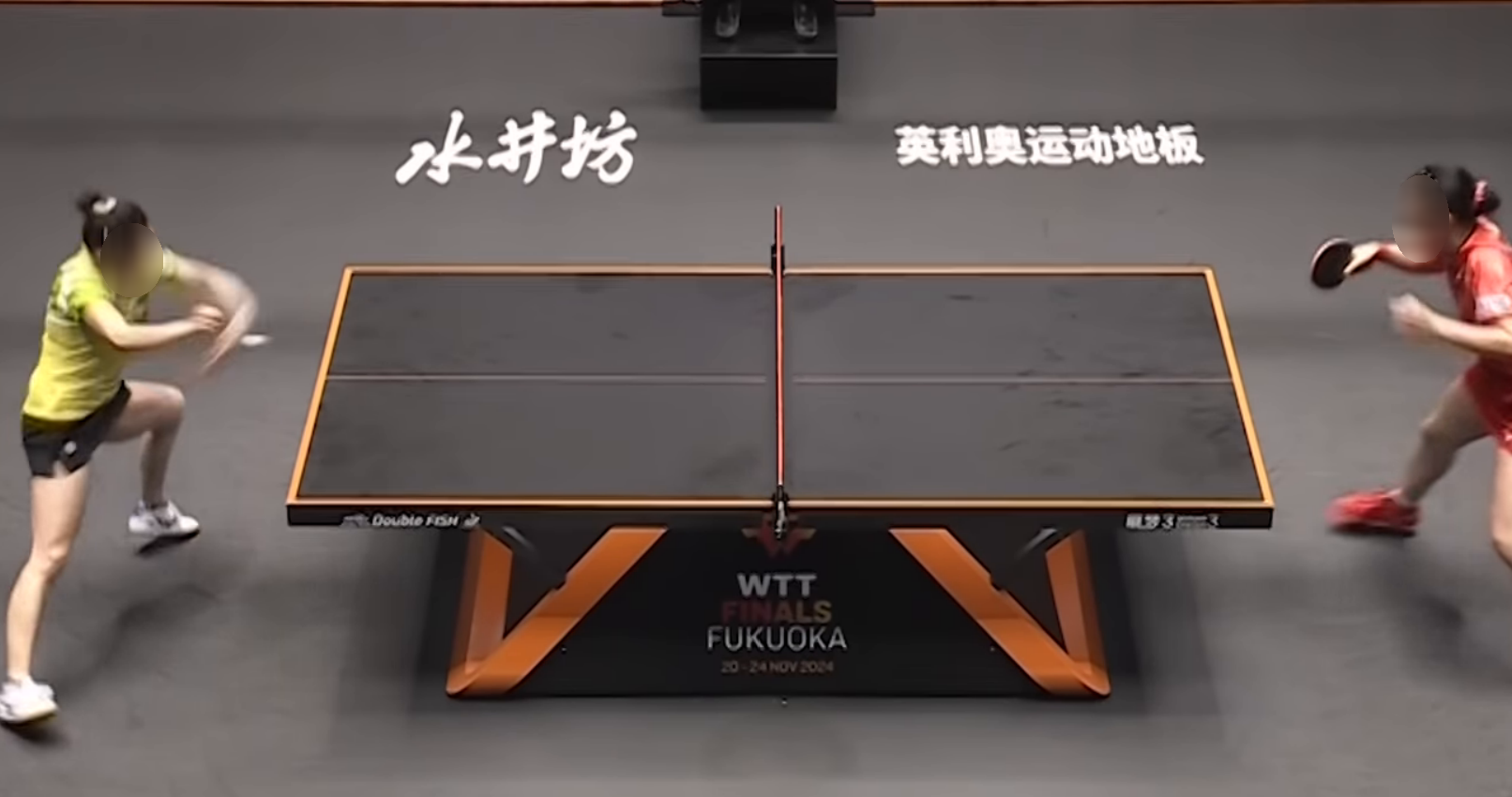}
        \caption{Topspin}
        \label{fig:topspin}
    \end{subfigure}
    \hfill
    \begin{subfigure}[b]{0.9\linewidth}
        \includegraphics[width=\linewidth]{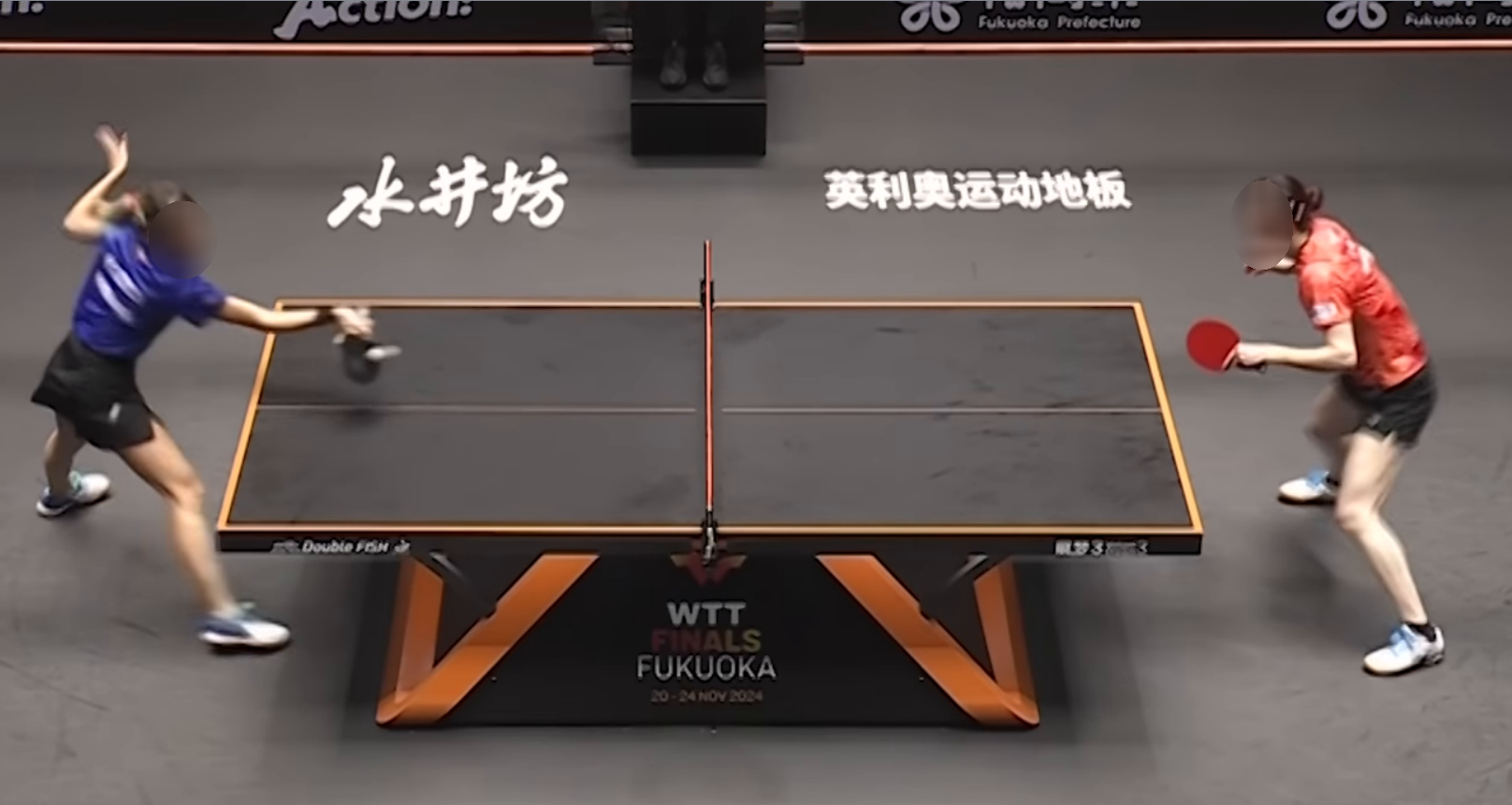}
        \caption{Backspin}
        \label{fig:backspin}
    \end{subfigure}
    \caption{Example frames for the annotation of the spin direction. If the paddle is facing the table (a), the shot is annotated as topspin. If the paddle is facing away from the table (b), the shot is annotated as backspin.}
    \label{fig:spinannotation}
\end{figure}

For each trajectory, we annotate the 13 table keypoints in the first frame. 
Since the camera remains static throughout each video, these annotations are consistently used for all subsequent frames within the trajectory. 
The annotated keypoints are illustrated in Figure \ref{img:tablekeypoints}.
\noindent Annotating the spin direction is particularly challenging, as the spin is not directly observable in broadcast footage. 
To infer the spin type, we analyze the paddle's orientation at the moment of impact.
\begin{itemize}
    \item \textbf{Topspin}: If the paddle is angled towards the table, the shot is labeled as topspin. This is illustrated in Figure \ref{fig:topspin}.
    \item \textbf{Backspin}: If the paddle is angled away from the table, the shot is labeled as backspin. This is illustrated in Figure \ref{fig:backspin}.
\end{itemize}
This annotation approach provides a practical method for inferring spin direction despite the limitations of broadcast video data.

\section{Regressing camera matrices}
\label{sec:regresscameramatrices}
Although our method operates in an end-to-end manner without requiring camera calibration as input, the intrinsic and extrinsic camera matrices are necessary for computing the 2D reprojection error.
For each trajectory, we manually annotate the 13 table keypoints once.
Since the corresponding 3D world coordinates are known due to standardized table sizes in professional matches, the camera matrices can be estimated by minimizing the reprojection error of these 13 points.
However, this regression process is inherently unstable, and even small annotation errors can lead to significant inaccuracies in the estimated camera matrices.
To mitigate this issue, we employ the RANSAC algorithm \cite{ransac} to robustly filter out erroneous annotations.
In each RANSAC iteration, we randomly select six non-planar keypoints and compute an initial estimate of the camera matrices using the Direct Linear Transformation (DLT) algorithm \cite{DLT}.
This initial estimate is then refined using the Broyden-Fletcher-Goldfarb-Shanno (BFGS) optimization algorithm \cite{BFGS}.
With the refined matrices, we determine the number of inliers by checking the reprojection error.
A point is classified as an inlier if the reprojected 3D point is within \SI{3}{pixels} of the corresponding 2D annotation.
This procedure is repeated 100 times, and the setting with the highest number of inliers is selected.
Using all identified inliers, we compute the final camera matrices by first applying the DLT algorithm and subsequently refining the result with the BFGS optimization.
The RANSAC-based approach is essential for mitigating the impact of slight errors in the 2D annotations and obtaining accurate camera matrices.
We highlight that our model, which does not require camera calibration, is very robust to minor errors in the 2D input, highlighting the benefits of implementing an end-to-end approach.

\section{Augmentation Details}
\label{sec:augmentationdetails}
\begin{figure}[h]
    \centering
    \begin{subfigure}[b]{0.49\linewidth}
        \includegraphics[width=\linewidth]{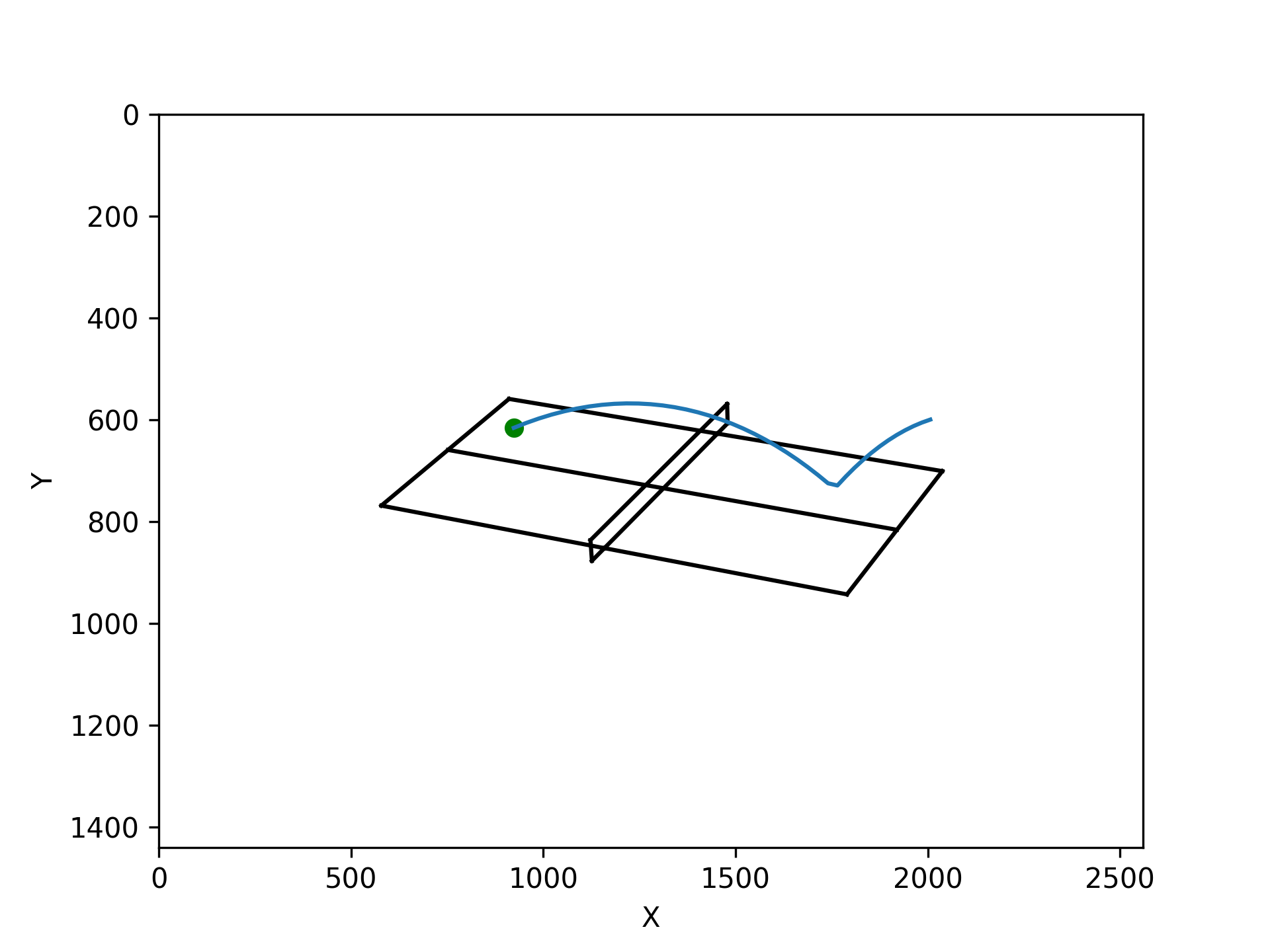}
    \end{subfigure}
    \hfill 
    \begin{subfigure}[b]{0.49\linewidth}
        \includegraphics[width=\linewidth]{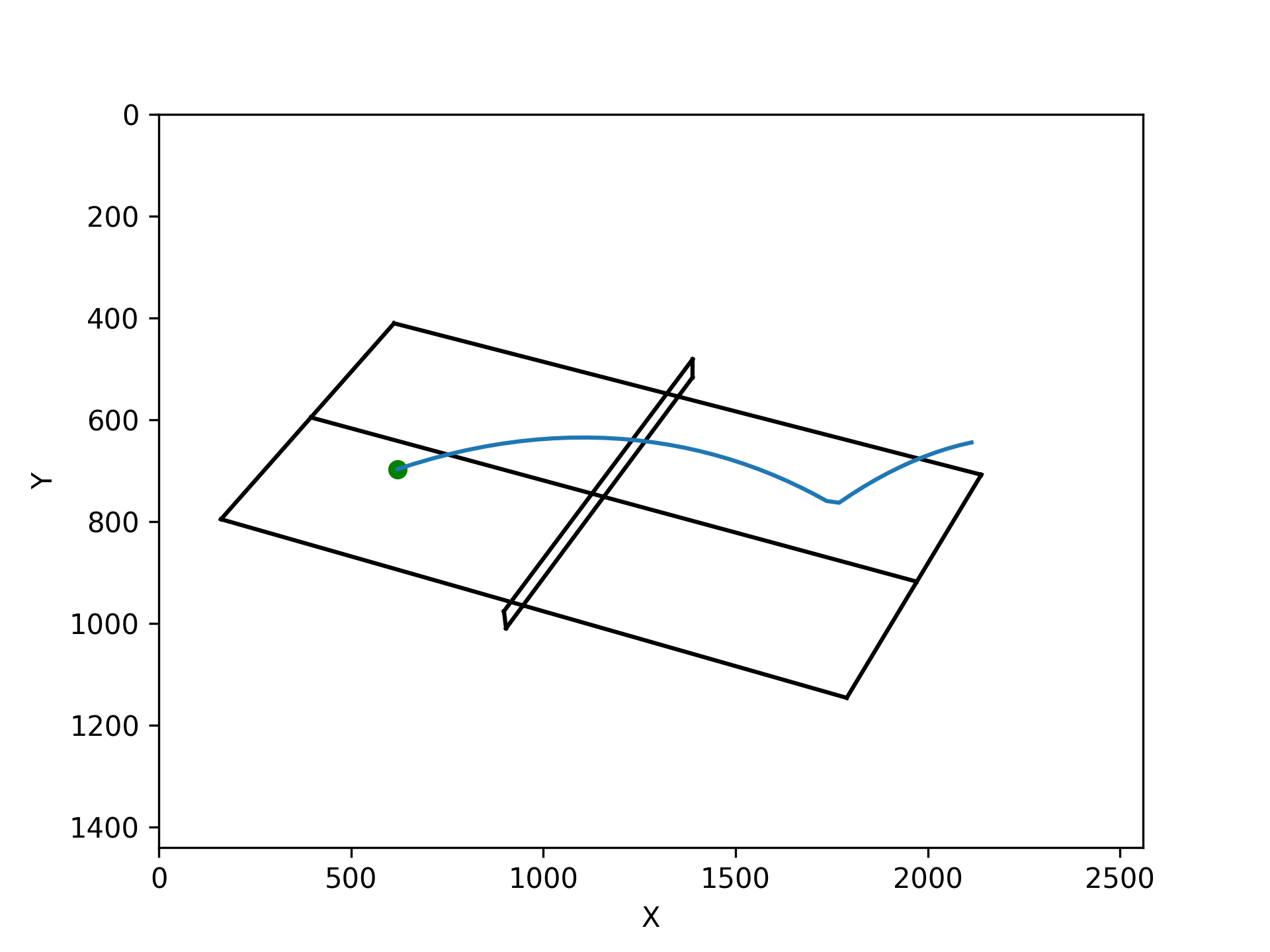}
    \end{subfigure}

    \begin{subfigure}[b]{0.49\linewidth}
        \includegraphics[width=\linewidth]{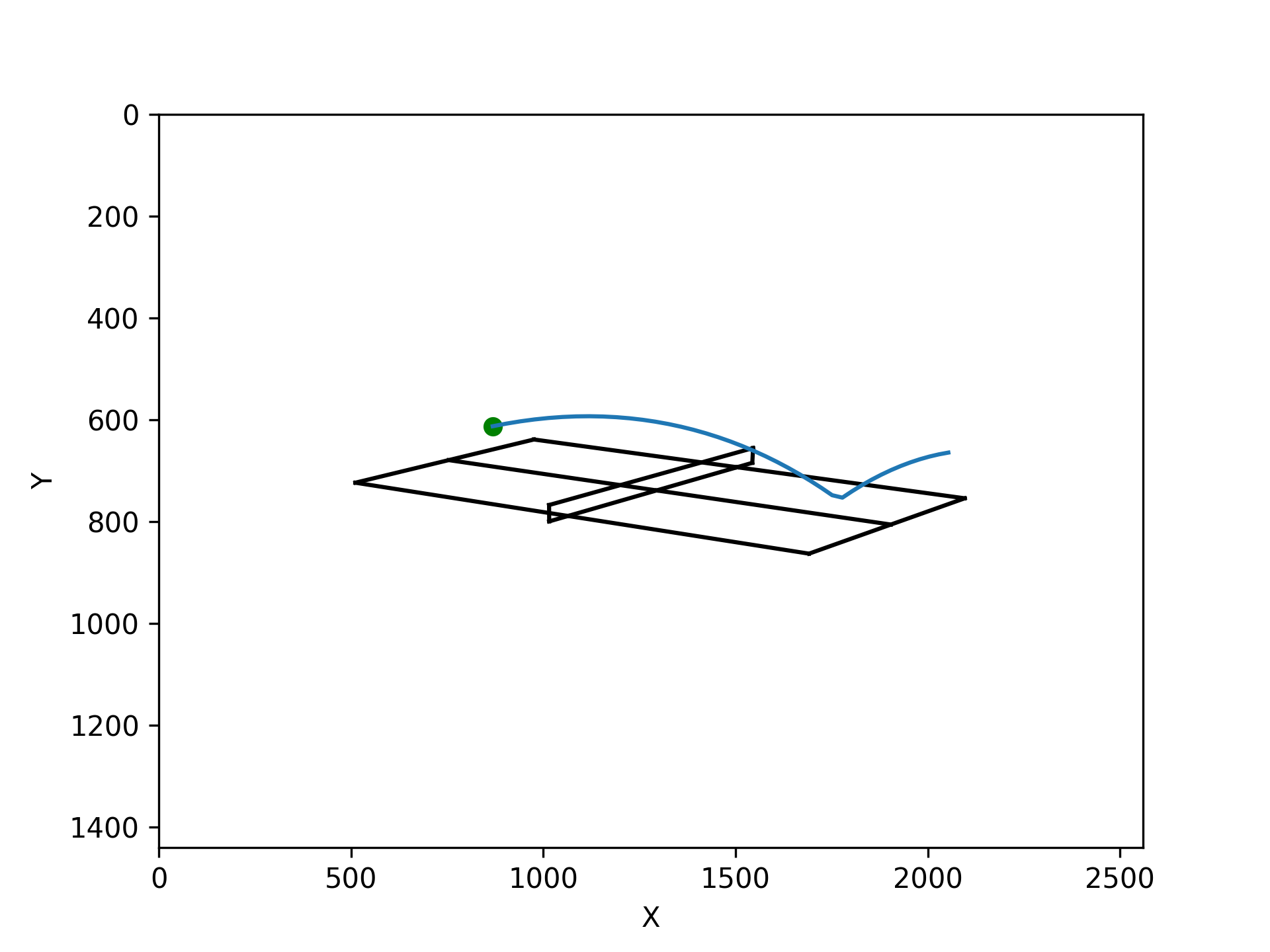}
    \end{subfigure}
    \hfill
    \begin{subfigure}[b]{0.49\linewidth}
        \includegraphics[width=\linewidth]{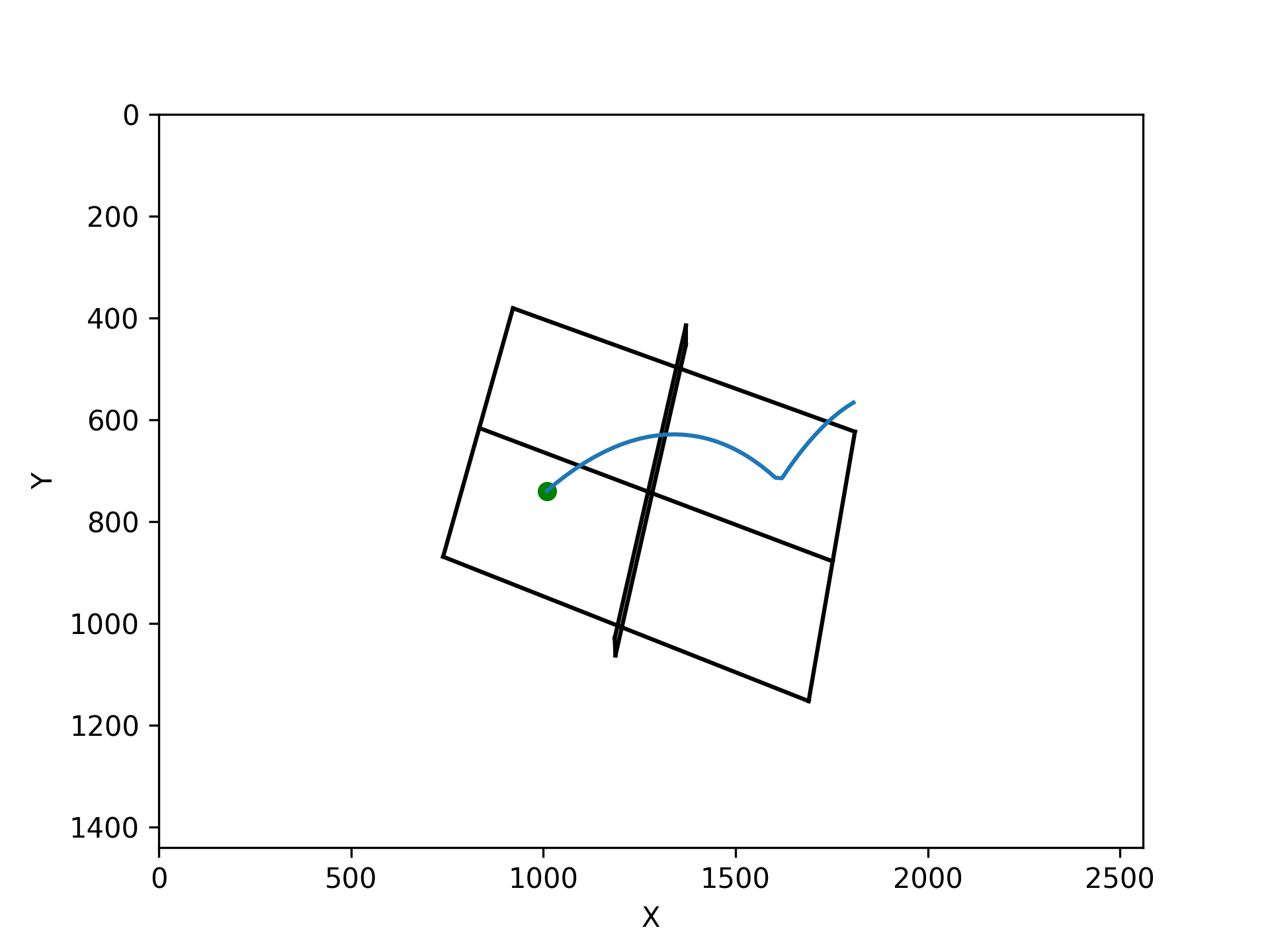}
    \end{subfigure}
    \caption{A trajectory sampled from 4 different camera perspectives. During training, we transform the trajectory with such randomly sampled camera parameters to introduce different viewpoints.}
    \label{fig:sampled_cameras}
\end{figure}
\noindent During training, we \textbf{randomly sample plausible camera parameters} to simulate diverse camera perspectives, increasing the diversity of the model inputs and ensuring that the model generalizes well across different viewpoints.
Examples of trajectories visualized from various sampled camera parameters are shown in Figure \ref{fig:sampled_cameras}. \\[0.5ex]
In the synthetic dataset, we do not only store the ball's 3D position at each time step but also record intermediate positions.
For these intermediate positions, we employ a "virtual" framerate of \SI{500}{Hz}, which is ten times higher than the actual framerate.
This allows us to accurately simulate \textbf{motion blur} in a physically consistent manner.
Rather than using the exact ball position at a given timestamp, we randomly select a point within a temporal window around each frame.
For the experiments in this paper, we define this window such that the selected point has a timestamp within a maximum deviation of $0.4*\frac{1}{\SI{50}{Hz}}$.
Motion blur is applied not only to the 2D ball positions but the 3D ground truth positions are shifted accordingly. \\[0.5ex]
The \textbf{sudden end} augmentation is designed to simulate scenarios in which the opposing player interferes, leading to an abrupt termination of the trajectory.
For each trajectory, we remove a randomly selected number of coordinates at the end, however, we ensure that the trajectory always includes the bounce on the table.
This augmentation is applied with a probability of $50\%$ during training, allowing the model to learn from complete trajectories while also adapting to cases where the ball's motion is unexpectedly interrupted. \\[0.5ex]
The \textbf{Gaussian blur} augmentation is implemented by introducing random noise to the 2D ball position and the 2D table keypoints.
The noise is sampled from a Gaussian distribution with a standard deviation of $2$ pixels in both the $x$- and $y$-directions, effectively simulating annotation inaccuracies.

\section{Further Experiments}
\label{sec:furtherexperiments}
This section presents additional experiments that provide deeper insights into the model's behavior.
All experiments are based on the best model, which is described in Section \ref{sec:detailedeval}.
It uses the concatenation token embedding method, the connect-stage SPT architecture, and all data augmentations.

\subsection{Spin Prediction Coordinate System}
\begin{table}[h]
    \centering
    \resizebox{\linewidth}{!}{ 
    \begin{tabular}{l|c|c|c|c|c|c}
        \toprule
         & \multicolumn{2}{c|}{Synthetic} & \multicolumn{4}{c}{Real} \\
        Method & $\Delta \vec{\omega}$ & $\Delta \vec{r}_\text{world}$ & $\textit{acc}$ & $\mathrm{F}_1$ & ROC-AUC & $\Delta \vec{r}_\text{img}$ \\
        \midrule
        world & \SI{48.7}{Hz} & \SI{5.5}{cm} & \SI{92.0}{\%} & \SI{0.917}{} & \SI{0.990}{} & \bfqty{0.19}{\%} \\
        ball & \SI{48.3}{Hz} & \SI{5.4}{cm} & \bfqty{94.0}{\%} & \bfqty{0.938}{} & \bfqty{1.000}{} & \SI{0.22}{\%} \\
        \bottomrule
    \end{tabular}
    }
    \caption{Comparison of different spin prediction coordinate systems. The best results on the real data are highlighted in bold.}
    \label{tab:results_spin}
\end{table}
\noindent In Section \ref{sec:coordsystems}, we discussed that while the trajectory is analyzed in the world coordinate system, the spin is evaluated in the ball coordinate system.
According to Equation \ref{eq:localcoords}, the predicted spin can be transformed between coordinate systems.
Thus, there are two approaches for predicting the spin:
\begin{itemize}
    \item The network is trained to predict the spin in the world coordinate system, and the predicted trajectory is used in Equation \ref{eq:localcoords} to transform the spin into the ball coordinate system.
    \item The network is trained to directly predict the spin in the ball coordinate system, eliminating the need for any transformation.
\end{itemize}
Since the first approach relies on the predicted trajectory for coordinate transformation, it may introduce additional errors.
Conversely, using the same coordinate system for both trajectory and spin could simplify training, as the network does not need to learn the transformation. \\[0.5ex]
Table \ref{tab:results_spin} compares both approaches.
Training the network directly in the ball coordinate system results in slightly better performance across all spin-related metrics.
However, trajectory prediction benefits from predicting the spin in the world coordinate system.
Overall, the differences are minor.
Choosing the coordinate system depends on whether trajectory accuracy or spin accuracy is more critical for the specific application.
This allows for flexibility in the model design, enabling it to be tailored to the specific requirements of the task at hand.

\subsection{Positional Encoding}
\begin{table}[h]
    \centering
    \resizebox{\linewidth}{!}{ 
    \begin{tabular}{l|c|c|c|c|c|c}
        \toprule
         & \multicolumn{2}{c|}{Synthetic} & \multicolumn{4}{c}{Real} \\
        Method & $\Delta \vec{\omega}$ $\downarrow$ & $\Delta \vec{r}_\text{world}$ $\downarrow$ & $\textit{acc}$ $\uparrow$ & $\mathrm{F}_1$ $\uparrow$ & ROC-AUC $\uparrow$ & $\Delta \vec{r}_\text{img}$ $\downarrow$ \\
        \midrule
        rotary & \SI{48.7}{Hz} & \SI{5.5}{cm} & \bfqty{92.0}{\%} & \bfqty{0.917}{} & \bfqty{0.990}{} & \bfqty{0.19}{\%} \\
        added & \SI{52.6}{Hz} & \SI{5.8}{cm} & \SI{90.0}{\%} & \SI{0.897}{} & \SI{0.987}{} & \SI{0.26}{\%} \\
        \bottomrule
    \end{tabular}
    }
    \caption{Comparison of different positional encodings. The best results on the real data are highlighted in bold.}
    \label{tab:results_positional}
\end{table}
\noindent The standard approach for incorporating positional information in transformers is by adding a fixed sinusoidal positional encoding to the token embeddings.
However, our model utilizes a rotary positional encoding, which is commonly used in language models.
Table \ref{tab:results_positional} compares both methods.
As the rotary positional encoding achieves better performance across all metrics, we conclude that it is more suitable for our task.

\subsection{Loss Target}
\begin{table}[h]
    \centering
    \resizebox{\linewidth}{!}{ 
    \begin{tabular}{c|c|c|c|c|c|c|c}
        \toprule
        \multicolumn{2}{c|}{Method} & \multicolumn{2}{c|}{Synthetic} & \multicolumn{4}{c}{Real} \\
        $\mathcal{L}_\text{trajectory}$ & $\mathcal{L}_\text{spin}$ & $\Delta \vec{\omega}$ $\downarrow$ & $\Delta \vec{r}_\text{world}$ $\downarrow$ & $\textit{acc}$ $\uparrow$ & $\mathrm{F}_1$ $\uparrow$ & ROC-AUC $\uparrow$ & $\Delta \vec{r}_\text{img}$ $\downarrow$ \\
        \midrule
        \ding{51} & \ding{51} & \SI{48.7}{Hz} & \SI{5.5}{cm} & \bfqty{92.0}{\%} & \bfqty{0.917}{} & \bfqty{0.990}{} & \bfqty{0.19}{\%} \\
        \ding{53} & \ding{51} & \SI{60.6}{Hz} & - & \SI{78.0}{\%} & \SI{0.769}{} & \SI{0.890}{} & - \\
        \ding{51} & \ding{53} & - & \SI{5.5}{cm} & - & - & - & \SI{0.22}{\%} \\
        \bottomrule
    \end{tabular}
    }
    \caption{Comparison of joint prediction with individual models for each task. \ding{51} indicates which loss function is used during training, while \ding{53} indicates the absence of the specific loss. The best results on the real data are highlighted in bold.}
    \label{tab:results_loss}
\end{table}
\noindent Our model is designed to jointly predict both trajectory and spin.
For training both task, we simply sum the two loss functions in Equation \ref{eq:loss}.
In this section, we examine whether joint prediction is beneficial or if the two tasks should be handled by separate models. \\[0.5ex]
Table \ref{tab:results_loss} compares joint prediction with separate models.
The results clearly show that joint prediction outperforms the separate models across all metrics.
This suggests that the network extracts useful information from one task that enhances the other.
Thus, joint prediction is an effective approach.

\subsection{Model Size}
\begin{table}[h]
    \centering
    \resizebox{\linewidth}{!}{ 
    \begin{tabular}{l|c|c|c|c|c|c}
        \toprule
         & \multicolumn{2}{c|}{Synthetic} & \multicolumn{4}{c}{Real} \\
        Method & $\Delta \vec{\omega}$ $\downarrow$ & $\Delta \vec{r}_\text{world}$ $\downarrow$ & $\textit{acc}$ $\uparrow$ & $\mathrm{F}_1$ $\uparrow$ & ROC-AUC $\uparrow$ & $\Delta \vec{r}_\text{img}$ $\downarrow$ \\
        \midrule
        small & \SI{64.9}{Hz} & \SI{10.7}{cm} & \bfqty{92.0}{\%} & \bfqty{0.917}{} & \SI{0.956}{} & \SI{0.29}{\%} \\
        base & \SI{51.0}{Hz} & \SI{6.2}{cm} & \SI{90.0}{\%} & \SI{0.895}{} & \bfqty{0.998}{} & \SI{0.25}{\%} \\
        large & \SI{48.7}{Hz} & \SI{5.5}{cm} & \bfqty{92.0}{\%} & \bfqty{0.917}{} & \SI{0.990}{} & \SI{0.19}{\%} \\
        huge & \SI{48.8}{Hz} & \SI{5.1}{cm} & \SI{86.0}{\%} & \SI{0.850}{} & \SI{0.971}{} & \bfqty{0.17}{\%} \\
        \bottomrule
    \end{tabular}
    }
    \caption{Comparison of different model sizes. The best results on the real data are highlighted in bold.}
    \label{tab:results_sizes}
\end{table}
\noindent Table \ref{tab:results_sizes} compares different model sizes, which are defined in Table \ref{tab:sizes}.
All models demonstrate good performance in spin prediction.
However, increasing the model size improves trajectory prediction accuracy.
Additionally, the spin prediction performance of the largest model is slightly worse than that of the other models, possibly due to overfitting.
Therefore, we identify the large model as the best compromise between spin prediction and trajectory prediction.

\section{Reproducibility and Open Resources}  
To facilitate reproducibility and further research, we provide the following resources:  
\begin{itemize}  
    \item Synthetic trajectories used for training.  
    \item Annotations for the real-world dataset.  
    \item Trained model weights.  
    \item Source code for both training and inference.  
\end{itemize}  
All resources are publicly available at \newline {\url{https://kiedani.github.io/CVPRW2025/}}.

\end{document}